\documentclass[10pt,twocolumn,letterpaper]{article}

\usepackage[pagenumbers]{cvpr}      
\makeatletter
\@namedef{ver@everyshi.sty}{}
\makeatother

\usepackage{graphicx}
\usepackage{amsmath}
\usepackage{amssymb}
\usepackage{booktabs}

\usepackage{tabularx}
\usepackage{multirow}
\usepackage{tikz}
\usepackage{arydshln}

\newcommand*\circled[1]{\tikz[baseline=(char.base)]{
            \node[shape=circle,draw,inner sep=1.0pt] (char) {#1};}}

\usepackage[pagebackref,breaklinks,colorlinks]{hyperref}

\usepackage[capitalize]{cleveref}
\crefname{section}{Sec.}{Secs.}
\Crefname{section}{Section}{Sections}
\Crefname{table}{Table}{Tables}
\crefname{table}{Tab.}{Tabs.}

\def\cvprPaperID{9881} 
\def\confName{CVPR}
\def\confYear{2023}

\begin{document}

\title{Efficient On-device Training via Gradient Filtering}


\author{Yuedong Yang\qquad Guihong Li\qquad Radu Marculescu\\
The University of Texas at Austin\\
{\tt\small \{albertyoung, lgh, radum\}@utexas.edu}
}
\maketitle

\begin{abstract}
Despite its importance for federated learning, continuous learning and many other applications,
on-device training remains an open problem for EdgeAI.
The problem stems from the large number of operations (\textit{e.g.}, floating point multiplications and additions) and memory consumption required during training by the back-propagation algorithm.
Consequently, in this paper, we propose a new gradient filtering approach which enables on-device CNN model training. More precisely, our approach creates a special structure with fewer unique elements in the gradient map, thus significantly reducing the computational complexity and memory consumption of back propagation during training.
Extensive experiments on image classification and semantic segmentation with multiple CNN models (\textit{e.g.}, MobileNet, DeepLabV3, UPerNet) and devices (\textit{e.g.}, Raspberry Pi and Jetson Nano) demonstrate the effectiveness and wide applicability of our approach. For example, compared to SOTA, we achieve up to 19$\times$ speedup and 77.1\% memory savings on ImageNet classification with only 0.1\% accuracy loss. Finally, our method is easy to implement and deploy; over 20$\times$ speedup and 90\% energy savings have been observed compared to highly optimized baselines in MKLDNN and CUDNN on NVIDIA Jetson Nano. Consequently, our approach opens up a new direction of research with a huge potential for on-device training.\footnote{Code: https://github.com/SLDGroup/GradientFilter-CVPR23}

\end{abstract}

\section{Introduction}
\label{sec:intro}

\begin{figure}[t]
    \centering
    \includegraphics[width=\linewidth]{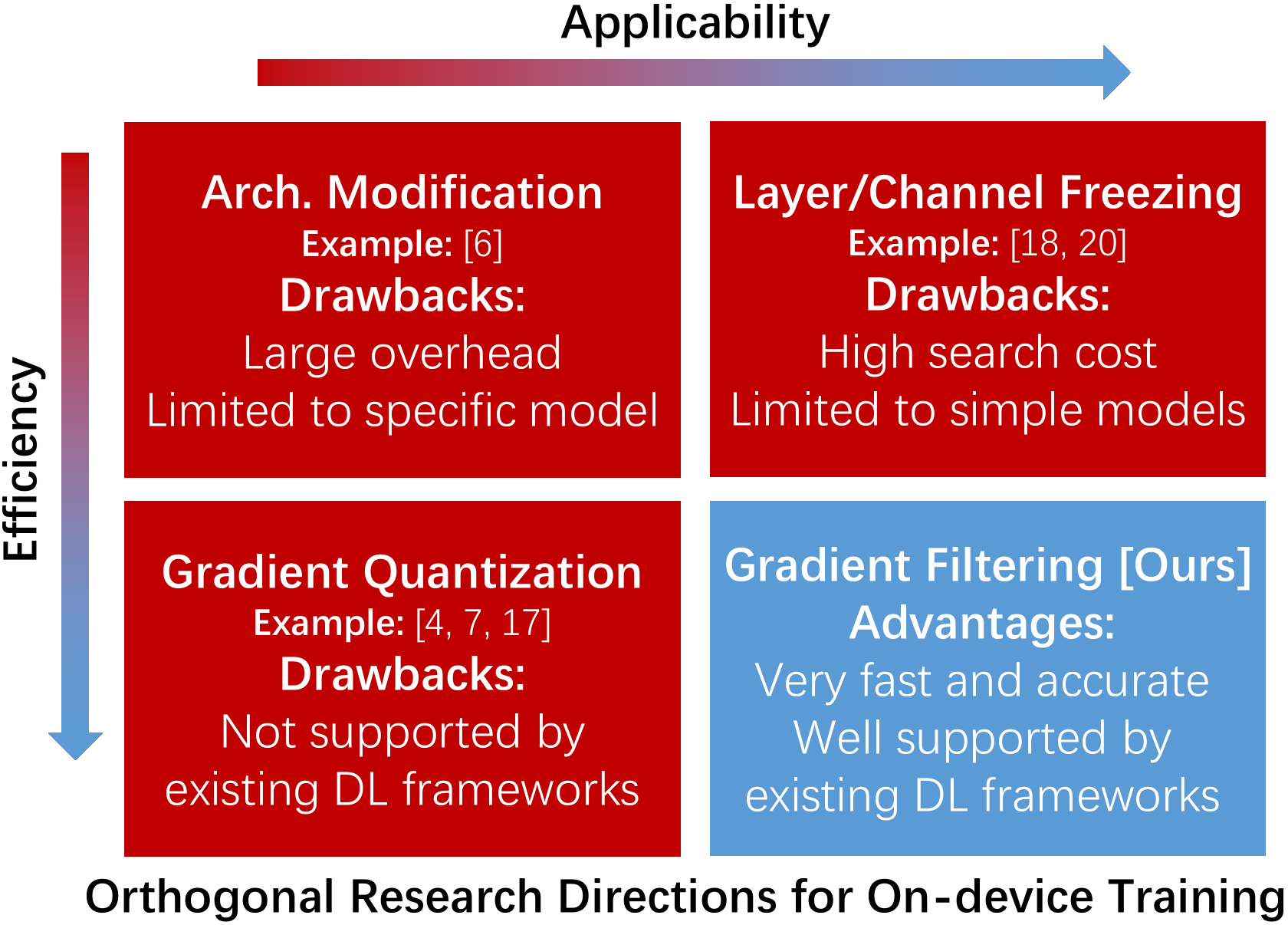}
    \caption{Matrix of orthogonal directions for on-device training. ``Arch'' is short for ``architecture''. Our approach opens up a new direction of research for on-device training for EdgeAI.}
    \label{fig:method_matrix}
\end{figure}

Existing approaches for on-device training are neither efficient nor practical enough to satisfy the resource constraints of edge devices (Figure \ref{fig:method_matrix}).
This is because these methods do not properly address a fundamental problem in on-device training, namely \textit{the computational and memory complexity of the back-propagation (BP) algorithm.}
More precisely, although the architecture modification~\cite{cai2020tinytl} and layer freezing~\cite{lin2022device, lee2022surgical} can help skipping the BP for some layers, for other layers, the complexity remains high. Gradient quantization~\cite{chen2020statistical, banner2018scalable} can reduce the cost of arithmetic operations but cannot reduce the number of operations (\textit{e.g.}, multiplications); thus, the speedup in training remains limited. Moreover, gradient quantization is not supported by existing deep-learning frameworks (e.g., CUDNN~\cite{chetlur2014cudnn}, MKLDNN~\cite{onednn}, PyTorch~\cite{pytorch} and Tensorflow~\cite{tensorflow2015-whitepaper}).
To enable on-device training, there are two important questions must be addressed:
\begin{itemize}
    \item \textit{How can we reduce the computational complexity of back propagation through the convolution layers?}
    \item \textit{How can we reduce the data required by the gradient computation during back propagation?}
\end{itemize}
In this paper, we propose \textit{gradient filtering}, a new research direction, to address both questions.
By addressing the first question, we reduce the computational complexity of training; by addressing the second question, we reduce the memory consumption.

In general, the gradient propagation through a convolution layer involves multiplying the gradient of the output variable with a Jacobian matrix constructed with data from either the input feature map or the convolution kernel.
We aim at simplifying this process with the new gradient filtering approach proposed in Section \ref{sec:method}.
Intuitively, if the gradient map w.r.t. the output has the same value for all entries, then the computation-intensive matrix multiplication can be greatly simplified, and the data required to construct the Jacobian matrix can be significantly reduced.
Thus, our gradient filtering can approximate the gradient w.r.t. the output by creating a new gradient map with a special (\textit{i.e.}, spatial) structure and fewer unique elements.
By doing so, the gradient propagation through the convolution layers reduces to cheaper operations, while the data required (hence memory) for the forward propagation also lessens. Through this filtering process, we trade off the gradient precision against the computation complexity during BP. We note that gradient filtering does not necessarily lead to a worse precision, \textit{i.e.}, models sometimes perform better with filtered gradients when compared against models trained with vanilla BP. 

In summary, our contributions are as follows:
\begin{itemize}
    \item We propose \textit{gradient filtering}, which reduces the computation and memory required for BP by more than two orders of magnitude compared to the exact gradient calculation.
    \item We provide a rigorous error analysis which shows that the errors introduced by the gradient filtering have only a limited influence on model accuracy.
    \item Our experiments with multiple CNN models and computer vision tasks show that we can train a neural network with significantly less computation and memory costs, with only a marginal accuracy loss compared to baseline methods. Side-by-side comparisons against other training acceleration techniques also suggest the effectiveness of our method.
    \item Our method is easy to deploy with highly optimized deep learning frameworks (\textit{e.g.}, MKLDNN~\cite{onednn} and CUDNN~\cite{chetlur2014cudnn}). Evaluations on resource-constrained edge (Raspberry Pi and Jetson Nano) and high-performance devices (CPU/GPU) show that our method is highly suitable for real life deployment.
\end{itemize}

The paper is organized as follows. Section \ref{sec:related_work} reviews relevant work. Section \ref{sec:method} presents our method in detail. Section \ref{sec:analysis} discusses error analysis, computation and memory consumption. Experimental results are presented in Section \ref{sec:exp}. Finally, Section \ref{sec:conclusion} summarizes our main contributions.

\section{Related Work}
\label{sec:related_work}
\noindent\textbf{Architecture Modification:} Authors of \cite{cai2020tinytl} propose to attach small branches to the original neural network. During training, the attached branches and biases in the original model are updated. Though memory consumption is reduced, updating these branches still needs gradient propagation through the entire network; moreover, a large computational overhead for inference is introduced.

\noindent\textbf{Layer Freezing:} Authors of \cite{lee2022surgical, lin2022device} propose to only train parts of the model. \cite{lee2022surgical} makes layer selection based on layer importance metrics, while \cite{lin2022device} uses evolutionary search. However, the layers selected by all these methods are typically computationally heavy layers (\textit{e.g.}, the last few layers in ResNet~\cite{he2016resnet}) which consume most of the resources. Thus, the speedup achieved by these approaches is limited.

\noindent\textbf{Gradient Quantization:} \cite{alistarh2016qsgd, bernstein2018signsgd} quantize gradient after back-propagation, which means these methods cannot accelerate the training on a single device. Work in \cite{chen2020statistical, banner2018scalable, sun2020ultra, hubara2017quantized, zhao2021distribution, hong2022efficient, wang2019e2} accelerates training by reducing the cost for every arithmetic operation. However, these methods do not reduce the number of operations, which is typically huge for SOTA CNNs, so their achievable speedup is limited. Also, all these methods are not supported by the popular deep learning frameworks~\cite{chetlur2014cudnn, onednn, tensorflow2015-whitepaper, pytorch}.

In contrast to the prior work, our method opens up a new research direction. More precisely, we reduce the number of computations and memory consumption required for training a single layer via gradient filtering. Thus, our method can be combined with any of the methods mentioned above. For example, in Section \ref{sec:supp_gq} in the Supplementary, we illustrate how our method can work together with the gradient quantization methods to enable a higher speedup.

%

\begin{figure*}[tp]
    \centering
    \includegraphics[width=\textwidth]{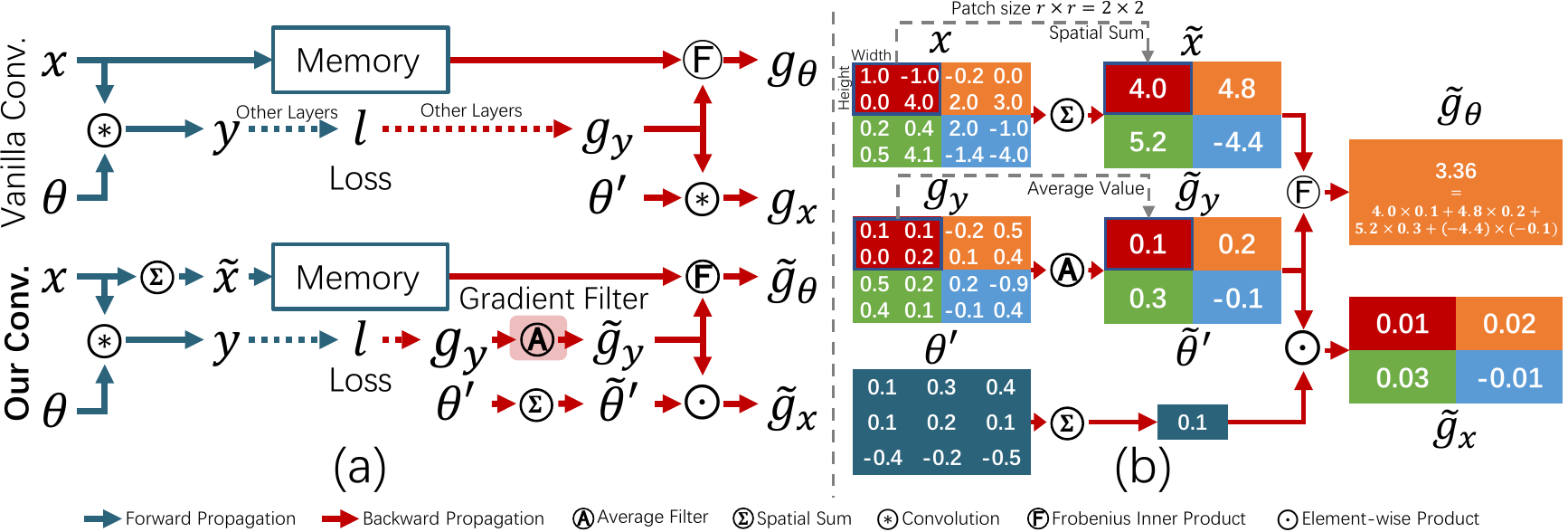}
    \caption{(a) Computation procedures for vanilla training method (upper) and our method (lower). (b) Example of gradient propagation with gradient filtering. Numbers in this example are chosen randomly for illustration purposes. In this case, the patch size selected for the gradient filter is $2\times 2$. Thus, the $4\times 4$ gradient map $g_y$ is approximated by $\tilde g_y$, which has four $2\times 2$ patches with one unique value for each patch. Also, input feature map $x$ and mirrored convolution kernel $\theta'$ are spatial summed to $\tilde x$ and $\tilde \theta'$. Since $\tilde x$ has fewer unique values than $x$, memory consumption is reduced. Finally, with $\tilde g_y, \tilde x$ and $\tilde \theta$, we compute the gradient w.r.t. kernel and input feature map with much fewer operations than the standard back propagation method.}
    \label{fig:method}
\end{figure*}

%

\begin{table}[b]
    \centering
    \begin{tabular}{c|c}
        \toprule
            $C_x$ & Number of channels of $x$ \\
            \hline
            $W_x, H_x$ & Width and height of $x$ \\
            \hline
            $\theta$ & Convolution kernel \\
            \hline
            $\theta'$ & Rotated $\theta$, \textit{i.e.}, $\theta'=\text{rot180}(\theta)$ \\
            \hline
            $r$ & Patch size ($r \times r$ ) \\
            \hline
            $g_x, g_y, g_\theta$ & Gradients w.r.t. $x, y, \theta$\\
            \hline
            $\tilde g_y$ & Approximated gradient $g_y$ \\
            \hline
            \multirow{2}{*}{$\tilde x, \tilde \theta'$} & Sum of $x$ and $\theta'$ over  \\
            ~ & spatial dimensions (height and width) \\
            \hline
            \multirow{2}{*}{$x[n, c_i, h, w]$} & Element for feature map $x$ \\
            ~ & at batch $n$, channel $c_i$, pixel $(h, w)$\\
            \hline
            \multirow{3}{*}{$\theta[c_o, c_i, u, v]$} & Element for convolution kernel $\theta$ \\
            ~ & at output channel $c_o$, input channel $c_i$,\\
            ~ & position $(u, v)$\\
        \bottomrule
    \end{tabular}
    \caption{Table of symbols we use.}
    \label{tab:symbols}
\end{table}

\section{Proposed Method}
\label{sec:method}

In this section, we introduce our gradient filtering approach to accelerate BP. To this end, we target the most computation and memory heavy operation, \textit{i.e.},  convolution (Figure \ref{fig:method}(a)). Table \ref{tab:symbols} lists some symbols we use.

\subsection{Problem Setup}

The computations for both forward and backward paths are shown in Figure~\ref{fig:method}(a).
For the standard (vanilla) approach (upper Figure~\ref{fig:method}(a)), starting with input $x$, the forward propagation convolves the input feature map $x$ with kernel $\theta$ and returns output $y$, which is further processed by the other layers in the neural network (dotted arrow) until the loss value $l$ is calculated. 
As shown in Figure \ref{fig:method}(a), the BP of the convolution layer starts with the gradient map w.r.t. output $y$ ($g_y$). The gradient w.r.t. input ($g_x$) is calculated by convolving $g_y$ with the \textit{rotated} convolution kernel $\theta'$, \textit{i.e.}, $g_x=g_y\circledast \text{rot180}(\theta)=g_y\circledast \theta'$. The gradient w.r.t. convolution kernel, namely $g_\theta$, is calculated with the Frobenius inner product \cite{horn2012matrix} between $x$ and $g_y$, \textit{i.e.}, $g_\theta=g_y\text{ \scriptsize\circled{F} }x$.

The lower half of Figure~\ref{fig:method}(a) shows our method, where several changes are made: 
We introduce the gradient filter ``{\scriptsize\circled{A}}'' after $g_y$ to generate the approximate gradient for BP. 
Also, instead of using the accurate $x$ and $\theta'$ values for gradient computation, we sum over spatial dimensions (height and width dimensions), \textit{i.e.}, $\tilde{x}$ and $\tilde{\theta}'$, respectively. Finally, the convolution layer now multiplies the approximate gradient $\tilde g_y$ with spatial kernel $\tilde\theta'$ instead of convolving with it to calculate $\tilde g_x$. Figure~\ref{fig:method}(b) shows an example of gradient propagation with our gradient filter.

\subsection{Preliminary Analysis}
\label{sec:preliminary_analysis}

Consider the vanilla BP for convolution in Figure \ref{fig:method}(a). Equation~\eqref{equ:vanilla_comp} shows the number of computations (\#FLOPs) required to calculate $g_x$ given $g_y$:

\begin{equation}
    \text{\#FLOPs} = 2C_{x}C_{y} \cdot W_y H_y \cdot W_{\theta}H_{\theta} 
    \label{equ:vanilla_comp}
\end{equation}
The computation requirements in Equation \eqref{equ:vanilla_comp} belong to three categories: number of channels, number of \textit{unique elements} per channel in the gradient map, and \textit{kernel size}. Our method focuses on the last two categories.

\textbf{i. Unique elements:} $(W_y H_y)$ represents the number of unique elements per channel in the gradient w.r.t. output variable $y$ ($g_y$). Given the high-resolution images we use, this term is huge, so if we manage to reduce the number of unique elements in the spatial dimensions (height and width), the computations required are greatly reduced too.

\textbf{ii. Kernel size:} $(W_{\theta}H_{\theta})$ represents the number of unique elements in the convolution kernel. If the gradient $g_y$ has some special structure, for example $g_y=1_{H_y\times W_y}\cdot v$ (\textit{i.e.}, every element in $g_y$ has the same value $v$), then the convolution can be simplified to $(\sum\theta')v1_{H_y\times W_y}$ (with boundary elements ignored). With such a special structure, only one multiplication and $(W_{\theta}H_{\theta}-1)$ additions are required. Moreover, $\sum\theta'$ is independent of data so the result can be shared across multiple images until $\theta$ gets updated.

\subsection{Gradient Filtering}
\label{sec:gf}
To reduce the number of unique elements and create the special structure in the gradient map, we apply the gradient filter after the gradient w.r.t. output ($g_y$) is provided. During the backward propagation, the gradient filter {\scriptsize\circled{A}} \textit{approximates} the gradient $g_y$ by spatially cutting the gradient map into $r\times r$-pixel patches and then replacing all elements in each patch with their average value (Figure \ref{fig:method}(b)):
\begin{equation}
    \tilde{g}_y[n, c_o, h, w] = \frac{1}{r^2}\sum_{i=\lfloor h/r\rfloor r}^{\lceil h/r\rceil r}\sum_{j=\lfloor w/r\rfloor r}^{\lceil w/r\rceil r} g_y[n, c_o, i, j]
\end{equation}
For instance in Figure \ref{fig:method}(b), we replace the 16 distinct values in the gradient map $g_y$ with 4 average values in $\tilde g_y$. 
So given a gradient map $g_y$ with $N$ images per batch, $C$ channels, and $H\times W$ pixels per channel, the gradient filter returns a structured approximation of the gradient map containing only $N\times C \times \lceil \frac{H}{r}\rceil \times \lceil\frac{W}{r}\rceil$ blocks, with \textit{one unique value per patch}. We use this matrix of unique values to represent the approximate gradient map $\tilde g_y$, as shown in Figure \ref{fig:method}(b).

\subsection{Back Propagation with Gradient Filtering}
\label{sec:bpgf}
We describe now the computation procedure used after applying the gradient filter. Detailed derivations are provided in Supplementary Section \ref{sec:supp_gf}.

\noindent\textbf{Gradient w.r.t. input:} The gradient w.r.t. input is calculated by convolving $\theta'$ with $g_y$ (Figure \ref{fig:method}(a)). With the approximate gradient $\tilde g_y$, this convolution simplifies to:
\begin{equation}
    \tilde g_{x}[n, c_i, h, w] = \sum_{c_o}\tilde g_{y}[n, c_o, h, w] \odot \tilde\theta'[c_o, c_i]
    \label{equ:grad_x}
\end{equation}
where $\tilde\theta'[c_o, c_i]=\sum_{u, v}\theta'[c_o, c_i, u, v]$ is the spatial sum of convolution kernel $\theta$, as shown in Figure~\ref{fig:method}(b).

\noindent\textbf{Gradient w.r.t. kernel:} The gradient w.r.t. the kernel is calculated by taking the Frobenius inner product between $x$ and $g_y$, \textit{i.e.}, $g_\theta[c_o, c_i, u, v] = x \text{ \scriptsize\circled{F} } g_y$, namely:
\begin{equation}
    g_\theta[c_o, c_i, u, v] = \sum_{n, i, j} x[n, c_i, i+u, j+v] g_y[n, c_o, i, j]
\end{equation}
With the approximate gradient $\tilde g_y$, the operation can be simplified to:
\begin{equation}
    \tilde g_{\theta}[c_o, c_i, u, v] = \sum_{n,i,j}\tilde x[n, c_i, i, j] \tilde g_{y}[n, c_o, i, j]
    \label{equ:grad_w}
\end{equation}
with $\tilde x[n, c_i, i, j] = \sum_{h=\lfloor i/r\rfloor r}^{\lceil i/r\rceil r}\sum_{w=\lfloor j/r\rfloor r}^{\lceil j/r\rceil r} x[n, c_i, h, w]$. As shown in Figure \ref{fig:method}(b), $\tilde x[n, c_i, i, j]$ is the spatial sum of $x$ elements in the same patch containing pixel $(i, j)$.

\section{Analyses of Proposed Approach}
\label{sec:analysis}
In this section, we analyze our method from three perspectives: gradient filtering approximation error, computation reduction, and memory cost reduction.
\subsection{Error Analysis of Gradient Filtering}
\label{sec:err_analysis}
We prove that the approximation error introduced by our gradient filtering is bounded during the gradient propagation. Without losing generality, we consider that all variables have only one channel, \textit{i.e.}, $C_{x_0}=C_{x_1}=1$. 

\noindent\textbf{Proposition 1:}
\textit{For any input-output channel pair $(c_o, c_i)$ in the convolution kernel $\theta$, assuming the DC component has the largest energy value compared to all components in the spectrum\footnote{As a reminder, the energy of a signal is the sum of energy of the DC component and the energy of its AC components.}, then the signal-to-noise-ratio (SNR) of $\tilde g_x$ is greater than SNR of $\tilde g_y$.}

\noindent\textbf{Proof:}
We use $G_x, G_y$ and $\Theta$ to denote the gradients $g_x, g_y$ and the convolution kernel $\theta$ in the \textit{frequency domain}; $G_x[u,v]$ is the spectrum value at frequency $(u,v)$ and $\delta$ is the 2D discrete Dirichlet function. To simplify the discussion, we consider only one patch of size $r \times r$. 

The gradient returned by the gradient filtering can be written as:
\begin{equation}
    \tilde g_y = \frac{1}{r^2} 1_{r\times r} \circledast g_y
    \label{equ:gf_spatial}
\end{equation}
where $\circledast$ denotes convolution. By applying the discrete Fourier transformation, Equation \eqref{equ:gf_spatial} can be rewritten in the frequency domain as:
\begin{equation}
\tilde G_y[u,v] = \frac{1}{r^2}\delta[u, v]G_y[u, v]
\label{equ:gf_freq}
\end{equation}
$\tilde g_y$ is the approximation of $g_y$ (\textit{i.e.}, the ground truth for $\tilde g_y$ is $g_y$), and the SNR of $\tilde g_y$ equals to:
\begin{equation}
\begin{aligned}
\text{SNR}_{\tilde g_y} 
&= \frac{\sum_{(u,v)}G_y^2[u,v]}{\sum_{(u,v)} (G_y[u,v]-\frac{1}{r^2}\delta[u,v]G_y[u,v])^2} \\
&= (1 - \frac{2r^2-1}{r^4}\frac{G_y^2[0,0]}{\sum_{(u,v)}G_y^2[u,v]})^{-1} \\
\end{aligned}
\label{equ:snr_g3}
\end{equation}
For the convolution layer, the gradient w.r.t. the approximate variable $\tilde x$ in the frequency domain is\footnote{Because $g_y$ is convolved with the \textbf{rotated} kernel $\theta'$, in the frequency domain, we use $\Theta[-u,-v]$ instead of $\Theta[u,v]$.}:
\begin{equation}
\begin{aligned}
	\tilde G_{x}[u,v] &= \Theta[-u,-v]\tilde G_{y}[u,v]\\ 
	&=\frac{1}{r^2}\Theta[-u,-v]\delta[u,v]G_y[u,v]
\end{aligned}
\end{equation}
and its ground truth is:
\begin{equation}
G_{x}[u,v] = \Theta[-u,-v]G_y[u,v]
\end{equation}
Similar to Equation \eqref{equ:snr_g3}, the SNR of $g_{\tilde x}$ is:
\begin{equation}
\begin{aligned}
    \text{SNR}_{\tilde g_x} = (1 - \frac{2r^2-1}{r^4}\frac{(\Theta[0,0]G_y[0,0])^2}{\sum_{(u,v)}{(\Theta[u,v]G_y[u,v])^2}})^{-1}
\end{aligned}
\label{equ:snr_approx_gx}
\end{equation}
Equation~\eqref{equ:snr_approx_gx} can be rewritten as:
\begin{equation}\label{equ:raw_prop}
    \begin{aligned}
        \frac{r^4(1-\text{SNR}^{-1}_{\tilde g_x})}{2r^2-1} &=\frac{(\Theta[0,0]G_y[0,0])^2}{\sum_{(u,v)} (\Theta[-u,-v]G_y[u,v])^2}\\
        &= \frac{G_y^2[0,0]}{\sum_{(u,v)} (\frac{\Theta[-u,-v]}{\Theta[0, 0]}G_y[u,v])^2}
    \end{aligned}
\end{equation}
Furthermore, the main assumption (\textit{i.e.}, the DC component dominates the frequency spectrum of $\Theta$) can be written as:
\begin{equation}\label{equ:dcac_ratio}
    \Theta^2[0,0]/\text{max}_{(u,v)\ne(0,0)} \Theta^2[u,v] \ge 1
\end{equation}
that is, $\forall (u, v), \frac{\Theta^2[-u,-v]}{\Theta^2[0,0]} \le 1$; thus, by combining Equation~\eqref{equ:raw_prop} and Equation~\eqref{equ:dcac_ratio}, we have:
\begin{equation}
\begin{aligned}
    \frac{G_y^2[0,0]}{\sum_{(u,v)} (\frac{\Theta[-u,-v]}{\Theta[0, 0]}G_y[u,v])^2} &\ge \frac{G_y^2[0,0]}{\sum_{(u,v)} (G_y[u,v])^2} \\ 
    \Leftrightarrow \frac{r^4(1-\text{SNR}^{-1}_{\tilde g_x})}{2r^2-1} &\ge \frac{r^4(1-\text{SNR}^{-1}_{\tilde g_y})}{2r^2-1}
\end{aligned}
\end{equation}
which means that: $\text{SNR}_{\tilde g_x} \ge \text{SNR}_{\tilde g_y}$. This completes our proof for error analysis. $\blacksquare$

In conclusion, as the gradient propagates through the network, the noise introduced by our gradient filter becomes weaker compared to the real gradient signal. This property ensures that the error in gradient has only a limited influence on the quality of BP.  We validate Proposition 1 later in the experimental section.

\subsection{Computation and Overhead Analysis}
\label{sec:comp_analysis}

\begin{figure}
    \centering
    \includegraphics[width=0.85\linewidth]{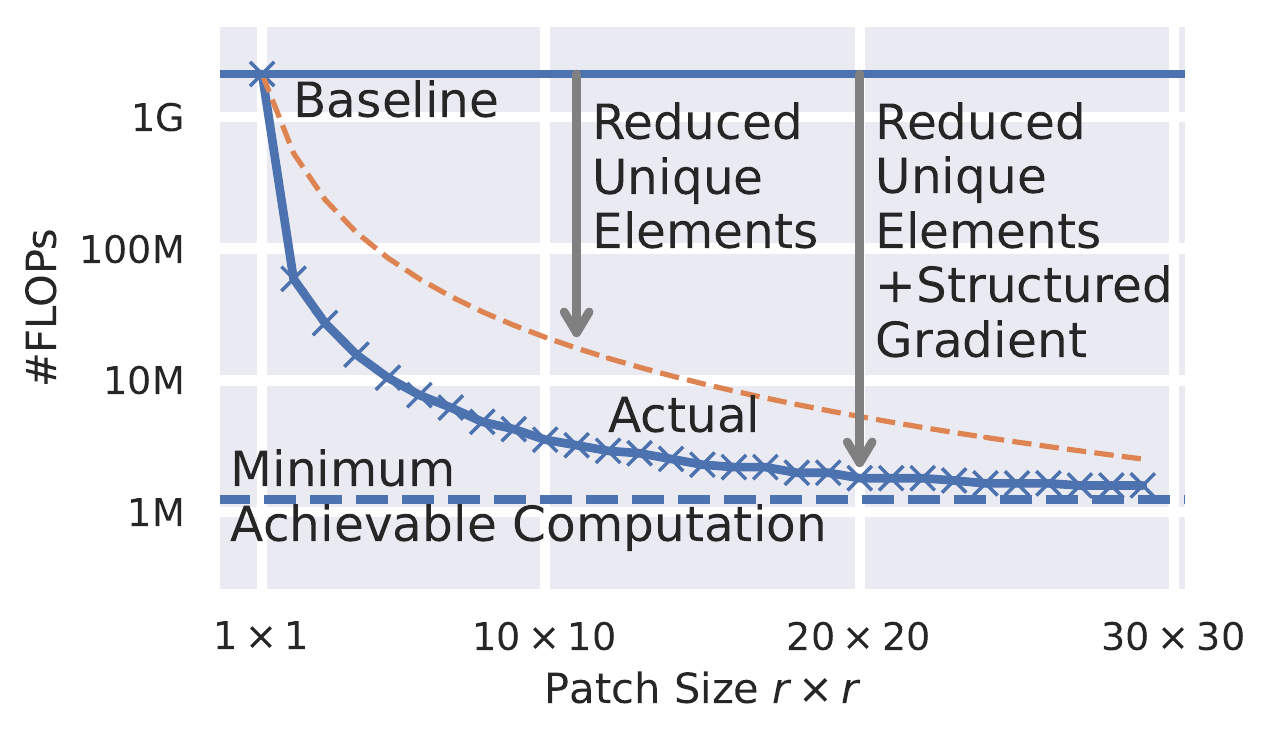}
    \caption{Computation analysis for a specific convolution layer\protect\footnotemark. Minimum achievable computation is given in Equation \eqref{equ:lb_comp}. By reducing the number of unique elements, computations required by our approach drop to about $1/r^2$ compared with the standard BP method. By combining it with structured gradient map, computations required by our approach drop further, getting very close to the theoretical limit.}
    \label{fig:comp_approx}
\end{figure}
\footnotetext{The layer is from U-Net~\cite{ronneberger2015unet}. The size of the input is assumed to be $120\times 160$ pixels with 192 channels; the output has the same resolution, but with only 64 channels. The kernel size of the convolution layer is $3\times3$. Analysis for ResNet is included in the supplementary material.}

In this section, we analyse the computation required to compute $g_x$, the gradient w.r.t. input $x$.
Figure \ref{fig:comp_approx} compares the computation required to propagate the gradient through this convolution layer under different patch sizes $r\times r$. A patch size $1\times 1$ means the vanilla BP algorithm which we use as the baseline. As discussed in the preliminary analysis section (Section \ref{sec:preliminary_analysis}), two terms contribute to the computation savings: fewer unique elements in the gradient map and the structured gradient map. 

\noindent\textbf{Fewer unique elements:} In vanilla BP, there are $H_y W_y$ unique elements in the gradient map. After applying gradient filtering with a patch size $r\times r$, the number of unique elements reduces to only $\lceil \frac{H_y}{r} \rceil \lceil\frac{W_y}{r}\rceil$. This reduction contributes the most to the savings in computation (orange line in Figure \ref{fig:comp_approx}).

\begin{table*}[htbp]
\centering
\resizebox{\textwidth}{!}{%
\begin{tabular}{cc|ccc|cc|ccc}
\toprule
\textbf{MobileNetV2}\cite{sandler2018mobilenetv2}                                                 & \textbf{\#Layers} & \textbf{Accuracy} & \textbf{FLOPs} & \textbf{Mem} & \textbf{ResNet-18}\cite{he2016resnet}                                                    & \textbf{\#Layers} & \textbf{Accuracy} & \textbf{FLOPs} & \textbf{Mem} \\ \hline
No Finetuning                                                         & 0                 & 4.2               & 0              & 0            & No Finetuning                                                         & 0                 & 4.7               & 0              & 0            \\ \hline
\multirow{3}{*}{\begin{tabular}[c]{@{}c@{}}Vanilla\\ BP\end{tabular}} & All               & 75.1              & 1.13G          & 24.33MB      & \multirow{3}{*}{\begin{tabular}[c]{@{}c@{}}Vanilla\\ BP\end{tabular}} & All               & 73.1              & 5.42G          & 8.33MB       \\ \cdashline{2-5}\cdashline{7-10}
                                                                      & 2                 & 63.1              & 113.68M        & 245.00KB     &                                                                       & 2                 & 70.4              & 489.20M        & 196.00KB     \\
                                                                      & 4                 & 62.2              & 160.00M        & 459.38KB     &                                                                       & 4                 & 72.3              & 1.14G          & 490.00KB     \\ \hline
TinyTL~\cite{cai2020tinytl}                                                                & N/A                 & 60.2                 & 663.51M        & 683.00KB     & TinyTL~\cite{cai2020tinytl}                                                                & N/A                 & 69.2                 & 3.88G          & 1.76MB       \\ \hline
\multirow{2}{*}{\textbf{Ours}}                                           & 2                 & 63.1              & 39.27M         & 80.00KB      & \multirow{2}{*}{\textbf{Ours}}                                           & 2                 & 68.6              & 28.32M         & 64.00KB      \\
                                                                      & 4                 & 63.4              & 53.96M         & 150.00KB     &                                                                       & 4                 & 68.5              & 61.53M         & 112.00KB     \\ \hline
\textbf{MCUNet}\cite{lin2020mcunet}                                                       & \textbf{\#Layers} & \textbf{Accuracy} & \textbf{FLOPs} & \textbf{Mem} & \textbf{ResNet-34}\cite{he2016resnet}                                                    & \textbf{\#Layers} & \textbf{Accuracy} & \textbf{FLOPs} & \textbf{Mem} \\ \hline
No Finetune                                                           & 0                 & 4.1               & 0              & 0            & No Finetune                                                           & 0                 &                   & 0              & 0            \\ \hline
\multirow{3}{*}{\begin{tabular}[c]{@{}c@{}}Vanilla\\ BP\end{tabular}} & All               & 68.5              & 231.67M        & 9.17MB       & \multirow{3}{*}{\begin{tabular}[c]{@{}c@{}}Vanilla\\ BP\end{tabular}} & All               & 70.8              & 11.17G         & 13.11MB      \\ \cdashline{2-5}\cdashline{7-10}
                                                                      & 2                 & 62.1              & 18.80M         & 220.50KB     &                                                                       & 2                 & 69.6              & 489.20M        & 196.00KB     \\
                                                                      & 4                 & 64.9              & 33.71M         & 312.38KB     &                                                                       & 4                 & 72.3              & 1.21G          & 392.00KB     \\ \hline
TinyTL~\cite{cai2020tinytl}                                                                & N/A                 & 53.1                 & 148.01M        & 571.5KB      & TinyTL~\cite{cai2020tinytl}                                                                & N/A                 & 72.9                 & 8.03G          & 2.95MB       \\ \hline
\multirow{2}{*}{\textbf{Ours}}                                           & 2                 & 61.8              & 6.34M          & 72.00KB      & \multirow{2}{*}{\textbf{Ours}}                                           & 2                 & 68.6              & 28.32M         & 64.00KB      \\
                                                                      & 4                 & 64.4              & 11.01M         & 102.00KB     &                                                                       & 4                 & 70.6              & 64.07M         & 128.00KB    
                                                                      \\ \bottomrule
\end{tabular}%
}
\caption{Experimental results for ImageNet classification with four neural networks (MobileNet-V2, ResNet18/34, MCUNet). ``\#Layers'' is short for ``the number of \textit{active} convolutional layers''. For example, \#Layers equals to 2 means that only the last two convolutional layers are trained. For memory consumption, we only consider the memory for input feature $x$. Strategy ``No Finetuning'' shows the accuracy on new datasets without finetuning the pretrained model. Since TinyTL~\cite{cai2020tinytl} changes the architecture, ``\#Layers'' is not applicable (N/A).}
\label{tab:imagenet}
\end{table*}

\begin{table*}[]
\centering
\resizebox{\textwidth}{!}{%
\begin{tabular}{cc|ccc|cc|ccc|cc|ccc}
\toprule
\textbf{PSPNet\cite{zhao2017pyramid}}                                                             & \textbf{\#Layers} & \textbf{GFLOPs} & \textbf{mIoU} & \textbf{mAcc} & \textbf{PSPNet-M\cite{zhao2017pyramid}}                                                           & \textbf{\#Layers} & \textbf{GFLOPs} & \textbf{mIoU} & \textbf{mAcc} & \textbf{FCN}\cite{long2015fully}                                                                & \textbf{\#Layers} & \textbf{GFLOPs} & \textbf{mIoU} & \textbf{mAcc} \\ \hline
Calibration                                                               & 0                 & 0               & 12.86         & 19.74         & Calibration                                                               & 0                 & 0               & 14.20         & 20.46         & Calibration                                                               & 0                 & 0               & 10.95         & 15.69         \\ \hline
\multirow{3}{*}{\begin{tabular}[c]{@{}c@{}}Vanilla\\ BP\end{tabular}}       & All               & 166.5           & 55.01         & 68.02         & \multirow{3}{*}{\begin{tabular}[c]{@{}c@{}}Vanilla\\ BP\end{tabular}}       & All               & 42.4           & 48.48         & 61.48         & \multirow{3}{*}{\begin{tabular}[c]{@{}c@{}}Vanilla\\ BP\end{tabular}}       & All               & 170.3           & 45.22         & 58.80         \\ \cdashline{2-5}\cdashline{7-10}\cdashline{12-15}
                                                                            & 5                 & 15.0            & 39.54         & 51.86         &                                                                             & 5                 & 12.22           & 36.35         & 47.09         &                                                                             & 5                 & 59.5            & 27.41         & 37.90         \\
                                                                            & 10                & 110.6           & 53.15         & 67.10         &                                                                             & 10                & 22.46           & 46.01         & 58.70         &                                                                             & 10                & 100.9           & 43.87         & 57.58         \\ \hline
\multirow{2}{*}{\textbf{Ours}} & 5                 & 0.14            & 39.34         & 51.86         & \multirow{2}{*}{\textbf{Ours}} & 5                 & 0.11            & 36.14         & 46.86         & \multirow{2}{*}{\textbf{Ours}} & 5                 & 0.58            & 27.42         & 37.88         \\
                                                                            & 10                & 0.79            & 50.88         & 64.73         &                                                                             & 10                & 0.76            & 44.90         & 57.50         &                                                                             & 10                & 0.96            & 36.30         & 48.82         \\ \hline
\textbf{DLV3}\cite{chen2017rethinking}                                                               & \textbf{\#Layers} & \textbf{GFLOPs} & \textbf{mIoU} & \textbf{mAcc} & \textbf{DLV3-M}\cite{chen2017rethinking}                                                             & \textbf{\#Layers} & \textbf{GFLOPs} & \textbf{mIoU} & \textbf{mAcc} & \textbf{UPerNet}\cite{xiao2018unified}                                                            & \textbf{\#Layers} & \textbf{GFLOPs} & \textbf{mIoU} & \textbf{mAcc} \\ \hline
Calibration                                                               & 0                 & 0               & 13.95         & 20.62         & Calibration                                                               & 0                 & 0               & 21.96         & 36.15         & Calibration                                                               & 0                 & 0               & 14.71         & 21.82         \\ \hline
\multirow{3}{*}{\begin{tabular}[c]{@{}c@{}}Vanilla\\ BP\end{tabular}}       & All               & 151.2           & 58.32         & 71.72         & \multirow{3}{*}{\begin{tabular}[c]{@{}c@{}}Vanilla\\ BP\end{tabular}}       & All               & 54.4            & 55.66         & 68.95         & \multirow{3}{*}{\begin{tabular}[c]{@{}c@{}}Vanilla\\ BP\end{tabular}}       & All               & 541.0           & 64.88         & 77.13         \\ \cdashline{2-5}\cdashline{7-10}\cdashline{12-15}
                                                                            & 5                 & 18.0            & 40.85            & 53.16             &                                                                             & 5                 & 14.8            & 38.21         & 49.35         &                                                                             & 5                 & 503.9           & 47.93         & 61.67         \\
                                                                            & 10                & 102.0           & 54.65         & 68.64         &                                                                             & 10                & 33.1            & 47.95         & 61.49         &                                                                             & 10                & 507.6           & 48.83             & 63.02             \\ \hline
\multirow{2}{*}{\textbf{Ours}} & 5                 & 0.31            & 33.09         & 44.33         & \multirow{2}{*}{\textbf{Ours}} & 5                 & 0.26            & 35.47         & 46.35         & \multirow{2}{*}{\textbf{Ours}} & 5                 & 1.97            & 47.04             & 60.44             \\
                                                                            & 10                & 2.96            & 47.11         & 60.28         &                                                                             & 10                & 1.40            & 45.53         & 58.99         &                                                                             & 10                & 2.22            & 48.00             & 62.07            
                                                                            \\ \bottomrule
\end{tabular}%
}
\caption{Experimental results for semantic segmentation task on augmented Pascal VOC12 dataset~\cite{chen2017rethinking}. Model name with postfix ``M'' means the model uses MobileNetV2 as backbone, otherwise ResNet18 is used. ``\#Layers'' is short for ``the number of \textit{active} convolutional layers'' that are trained. All models are pretrained on Cityscapes dataset~\cite{Cordts2016Cityscapes}. Strategy ``Calibration'' shows the accuracy when only the classifier and normalization statistics are updated to adapt different numbers of classes between augmented Pascal VOC12 and Cityscapes.}
\label{tab:seg}
\end{table*}

\noindent\textbf{Structured Gradient Map:} By creating the structured gradient map, the convolution over the gradient map $\tilde g_y$ is simplified to the element-wise multiplication and channel-wise addition. Computation is thus reduced to $(H_{\theta}W_{\theta})^{-1}$ of its original value. For instance, the example convolution layer in Figure \ref{fig:comp_approx} uses a $3\times3$ convolution kernel so around $89\%$ computations are removed. The blue line in Figure \ref{fig:comp_approx} shows the \#FLOPs after combining both methods. Greater reduction is expected when applying our method with larger convolution kernels. For instance, FastDepth~\cite{wofk2019fastdepth} uses $5\times5$ convolution kernel so as much as $96\%$ reduction in computation can be achieved, in principle.

\noindent\textbf{Minimum Achievable Computation:} With the two reductions mentioned above, the computation required to propagate the gradient through the convolution layer is:
\begin{equation}
    \text{\#FLOPs}(r) = \lceil \frac{H_{y}}{r} \rceil \lceil\frac{W_{y}}{r}\rceil C_{x}(2C_{y}-1) + o(H_{y}W_{y})
    \label{equ:min_comp}
\end{equation}
where $o(H_{y}W_{y})$ is a constant term which is independent of $r$ and negligible compared to $H_y W_y$. When the patch is as large as the feature map, our method reaches the  minimum achievable computation (blue dashed line in Figure \ref{fig:comp_approx}):
\begin{equation}
    \text{min}_r~\text{\#FLOPs}(r)=2C_{x}C_{y} - C_{x} + o(H_{y}W_{y})
\label{equ:lb_comp}
\end{equation}
In this case, each channel of the gradient map is represented with a single value, so the computation is controlled by the number of input and output channels. 

\noindent\textbf{Overhead:} The overhead of our approach comes from approximating the feature map $x$, gradient $g_y$, and kernel $\theta$. As the lower part of Figure \ref{fig:method}(a) shows, the approximation for $x$ is considered as part of forward propagation, while the other two as back propagation. Indeed, with the patch size $r$, the ratio of forward propagation overhead is about $1/(2C_o W_\theta H_\theta)$, while the ratio of backward propagation overhead is about $(r^2-1)/(2C_x)$.

Given the large number of channels and spatial dimensions in typical neural networks, both overhead values take less than 1\% computation in the U-Net example above.
\subsection{Memory Analysis}

As Figure \ref{fig:method}(a) shows, the standard back propagation for a convolution layer relies on the input feature map $x$, which needs to be stored in memory during forward propagation. Since every convolution layer requiring gradient for its kernel needs to save a copy of feature map $x$, the memory consumption for storing $x$ is huge. With our method, we simplify the feature map $x$ to approximated $\tilde x$, which has only $\lceil\frac{H_x}{r}\rceil\lceil\frac{W_x}{r}\rceil$ unique elements for every channel. Thus, by saving only these unique values, our method achieves around $(1 - \frac{1}{r^2})$ memory savings, overall.

\section{Experiments}
\label{sec:exp}

Our experimental section consists of theoretical and practical evaluations. Sections \ref{sec:exp_cls}-\ref{sec:exp_hyper} show the theoretical advantages of our method on image classification and semantic segmentation tasks with implementation-agnostic metrics (\textit{e.g.}, accuracy, FLOPs). Then, in Section \ref{sec:exp_ondevice}, we show how these theoretical advantages translate into practical advantages (\textit{i.e.,} speedup and memory savings) on real edge devices.


\subsection{Experimental Setup}
\label{sec:setup}
\noindent\textbf{Classification:} Following \cite{mcmahan2017communication}, we split every dataset into two highly non-i.i.d. partitions with the same size. Then, we pretrain our models on the first partition with a vanilla training strategy, and finetune the model on the other partition with different configurations for the training strategy (\textit{i.e.}, with/without gradient filtering, hyper-parameters, number of convolution layers to be trained).
More details (\textit{e.g.}, hyper-parameters) are in the Supplementary.

\noindent\textbf{Segmentation:} Models are pretrained on Cityscapes~\cite{Cordts2016Cityscapes} by MMSegmentation~\cite{mmseg2020}. Then, we calibrate and finetune these models with different training strategies on the augmented Pascal-VOC12 dataset following \cite{chen2017rethinking}, which is the combination of Pascal-VOC12~\cite{voc12} and SBD~\cite{sbd}.
More details are included in the supplementary material.

\noindent\textbf{On-device Performance Evaluation:}
For CPU performance evaluation, we implement our method with MKLDNN~\cite{onednn} (a.k.a. OneDNN) v2.6.0 and compare it with the convolution BP method provided by MKLDNN. We test on three CPUs, namely Intel 11900KF, Quad-core Cortex-A72 (Jetson Nano) and Quad-core Cortex-A53 (Raspberry Pi-3b). For GPU performance evaluation, we implement our method on CUDNN v8.2 \cite{chetlur2014cudnn} and compare with the BP method provided by CUDNN. We test on two GPUs, RTX 3090Ti and the edge GPU on Jetson Nano. Since both MKLDNN and CUDNN only support float32 BP, we test float32 BP only. Additionally, for the experiments on Jetson Nano, we record the energy consumption for CPU and GPU with the embedded power meter. More details (\textit{e.g.}, frequency) are included in the supplementary material.

\subsection{ImageNet Classification}
\label{sec:exp_cls}

Table \ref{tab:imagenet} shows our evaluation results on the ImageNet classification task. As shown, our method significantly reduces the FLOPs and memory required for BP, with very little accuracy loss. For example, for ResNet34, our method achieves 18.9$\times$ speedup with 1.7\% accuracy loss when training four layers; for MobileNetV2, we get a 1.2\% better accuracy with 3.0$\times$ speedup and 3.1$\times$ memory savings. These results illustrate the effectiveness of our method. On most networks, TinyTL has a lower accuracy while consuming more resources compared to the baselines methods.

\subsection{Semantic Segmentation}
\label{sec:exp_seg}

Table~\ref{tab:seg} shows our evaluation results on the augmented Pascal-VOC12 dataset. On a wide range of networks, our method constantly achieves significant speedup with marginal accuracy loss.
For the large network UPerNet, our method achieves 229$\times$ speedup with only 1\% mIoU loss. For the small network PSPNet, our method speedups training by 140$\times$ with only 2.27\% mIoU loss. This shows the effectiveness of our method on a dense prediction task.

\subsection{Hyper-Parameter Selection}
\label{sec:exp_hyper}
Figure \ref{fig:hyper_parameter} shows our experimental results for ResNets under different hyper-parameter selection, \textit{i.e.} number of convolution layers and patch size of gradient filter $r\times r$. Of note, the y-axis (MFLOPs) in Figure \ref{fig:hyper_parameter} is log scale. More results are included in Supplementary Section~\ref{sec:supp_hyper}. 
We highlight three qualitative findings in Figure \ref{fig:hyper_parameter}:

\begin{figure}[htbp]
    \centering
    \includegraphics[width=\linewidth]{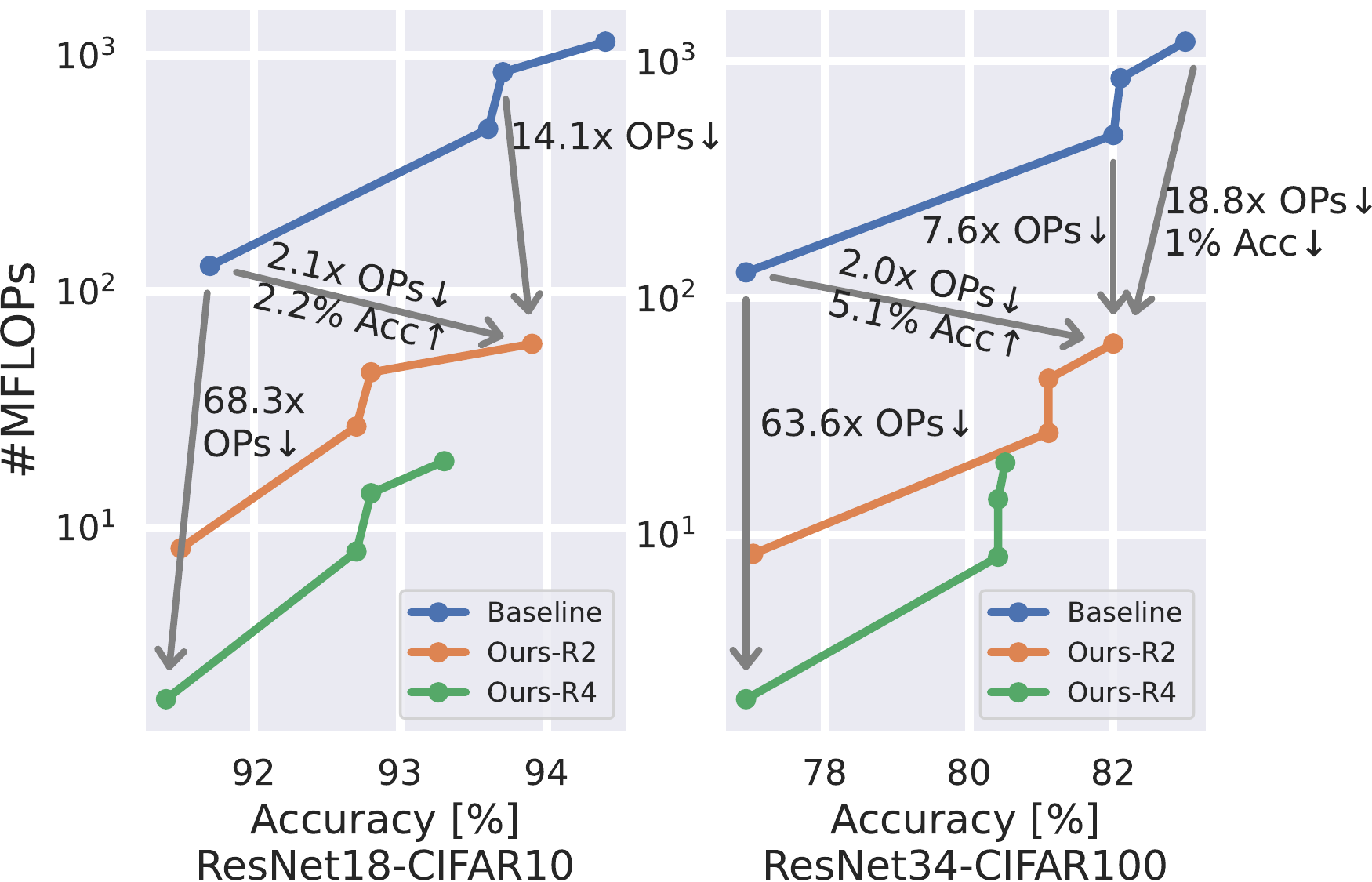}
    \caption{Computation (\#MFLOPs, log scale) and model accuracy [\%] under different hyper-parameter selection. ``Baseline'' means vanilla BP; ``Ours-R2/4'' uses gradient filtering with patch size $2\times 2$/$4\times 4$ during BP.}
    \label{fig:hyper_parameter}
\end{figure}
\begin{enumerate}
    \item[\textbf{a.}] For a similar accuracy, our method greatly reduces the number of operations (1 to 2 orders of magnitude), while for a similar number of computations, our method achieves a higher accuracy (2\% to 5\% better).
\end{enumerate}
This finding proves the effectiveness of our method.

\begin{figure}[h]
    \centering
    \includegraphics[width=0.8\linewidth]{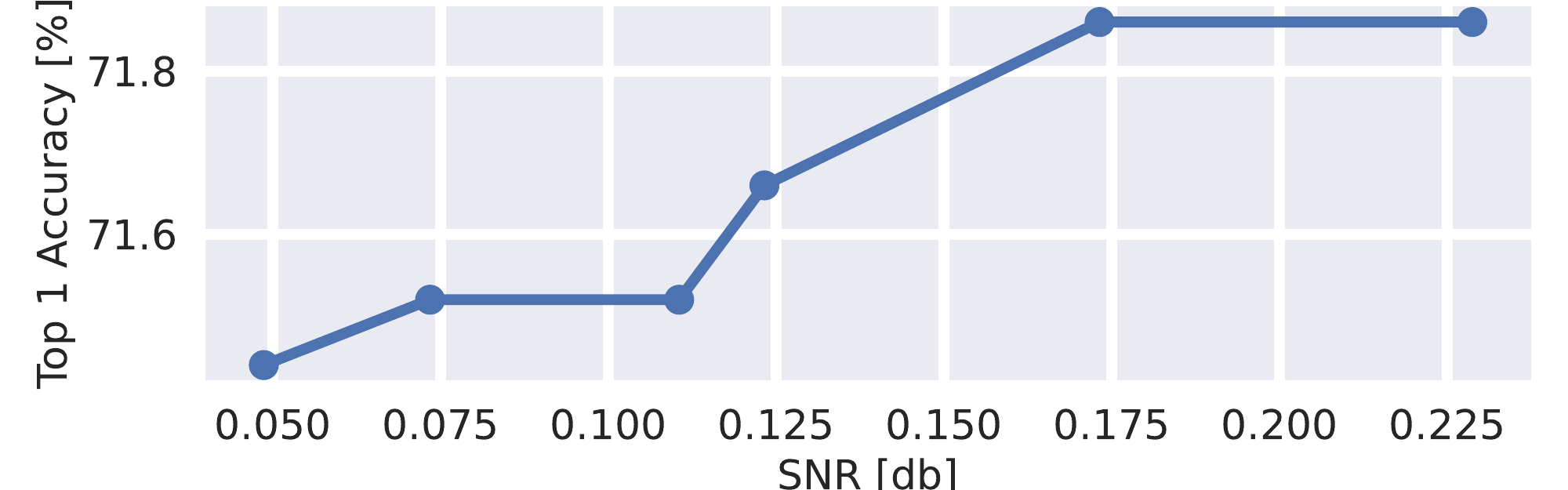}
    \caption{Relationship between accuracy and noise level introduced by the gradient filtering. As shown, accuracy increases as the SNR increases, \textit{i.e.}, noise level decreases.}
    \label{fig:snr}
\end{figure}

\begin{enumerate}
    \item[\textbf{b.}] Given the number of convolution layers to be trained, the more accurate method returns a better accuracy. Baseline (\textit{i.e.}, standard BP) uses the most accurate gradient, Ours-R4 (BP with gradient filter with patch size $4\times4$) uses the least accurate gradient; thus, Baseline $>$ Ours-R2 $>$ Ours-R4.
\end{enumerate}
This finding is intuitive since the more accurate method should introduce smaller noise to the BP, \textit{e.g.}, the gradient filtering with patch size $2\times 2$ (Ours-R2) introduces less noise than with patch size $4\times 4$ (Ours-R4). In Figure~\ref{fig:snr}, we evaluate the relationship between accuracy and noise level introduced by gradient filtering. With a higher SNR (\textit{i.e.}, a lower noise level), a better accuracy is achieved.
\begin{enumerate}
    \item[\textbf{c.}] Given the number of computations, the less accurate method returns the better accuracy by training more layers, \textit{i.e.}, Ours-R4 $>$ Ours-R2 $>$ baseline.
\end{enumerate}
This finding suggests that for neural network training with relatively low computational resources, training more layers with less accurate gradients is preferable than training fewer layers with more accurate gradients.

\begin{figure*}[htbp]
    \centering
    \includegraphics[width=0.95\linewidth]{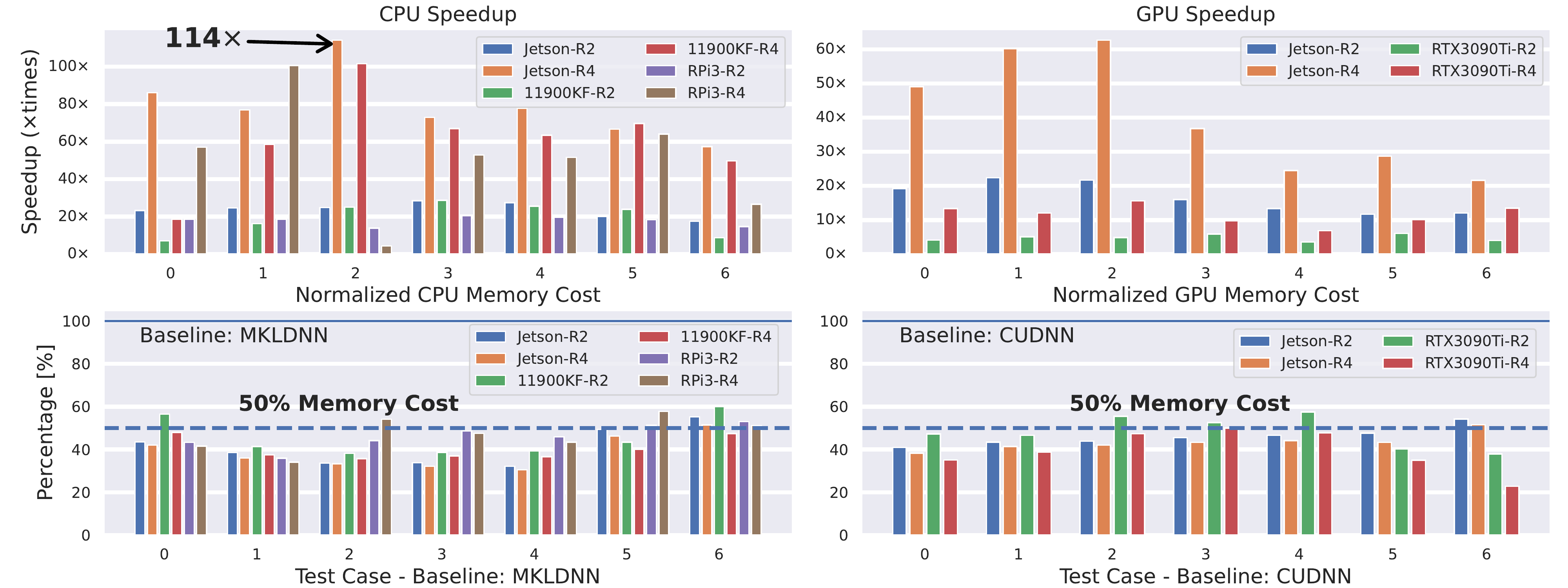}
    \caption{Speedup and normalized memory consumption results on multiple CPUs and GPUs under different test cases (\textit{i.e.} different input sizes, numbers of channels, etc.) Detailed configuration of these test cases are included in the supplementary material. ``R2'', ``R4'' mean using gradient filtering with $2\times 2$ and $4\times 4$ patch sizes, respectively. Our method achieves significant speedup with low memory consumption compared to all baseline methods. For example, on Jetson CPU with patch size $4\times 4$ (``Jetson-R4'' in left top figure), our method achieves 114$\times$ speedup with only 33\% memory consumption for most test cases.}
    \label{fig:speedup_mem}
\end{figure*}

\subsection{On-device Performance Evaluation}
\label{sec:exp_ondevice}

Figure \ref{fig:speedup_mem} and Table \ref{tab:energy} show our evaluation results on real devices. More results are included in the Supplementary Section~\ref{sec:supp_ondevice}. As Figure~\ref{fig:speedup_mem} shows, on CPU, most convolution layers achieve speedups over 20$\times$ with less than 50\% memory consumption for gradient filtering with patch sizes $2\times 2$; for gradient filtering with patch size $4\times 4$, the speedups are much higher, namely over 60$\times$. On GPU, the speedup is a little bit lower, but still over 10$\times$ and 25$\times$, respectively. Furthermore, as Table \ref{tab:energy} shows, our method saves over 95\% energy for both CPU and GPU scenarios, which largely resolves one of the most important constraints on edge devices. All these experiments on real devices show that our method is practical for the real deployment of both high-performance and IoT applications.

\begin{table}[]
\centering
\begin{tabular}{cc|c}
\toprule
Device                                                              & Patch Size & Normalized Energy Cost {[}STD{]} \\ \hline
\multirow{2}{*}{\begin{tabular}[c]{@{}c@{}}Edge\\ CPU\end{tabular}} & $2\times 2$          & 4.13\% {[}0.61\%{]}              \\
                                                                    & $4\times 4$          & 1.15\% {[}0.18\%{]}              \\ \hline
\multirow{2}{*}{\begin{tabular}[c]{@{}c@{}}Edge\\ GPU\end{tabular}} & $2\times 2$          & 3.80\% {[}0.73\%{]}              \\
                                                                    & $4\times 4$          & 1.22\% {[}1.10\%{]}             
                                                                    \\ \bottomrule
\end{tabular}
\caption{Normalized energy consumption for BP with gradient filtering for different patch sizes. Results are normalized w.r.t. the energy cost of standard BP methods. For instance, for edge CPU with a $4\times 4$ patch, only 1.15\% of energy in standard BP is used. Standard deviations are shown within brackets.}
\label{tab:energy}
\end{table}

\begin{table}[htbp]
\centering
\resizebox{\linewidth}{!}{
\begin{tabular}[t]{lc|lc}
   \toprule
   Model & Ratio & Model & Ratio \\
   \midrule
   (Wide)ResNet18-152 & 1.462 & VGG(bn)11-19 & 1.497 \\
   DenseNet121-201 & 2.278 & EfficientNet b0-b7 & 1.240 \\
   \bottomrule
\end{tabular}}
\caption{Evaluation of energy ratio defined in Equation~\eqref{equ:dcac_ratio} on models published on Torchvision. The ratio greater than 1 empirically verifies our assumption.}
\label{tab:dcac}
\end{table}

\subsection{Main Assumption Verification}
We now empirically verify the assumption that the DC component dominates the frequency spectrum of the convolution kernel (Section~\ref{sec:err_analysis}). To this end, we collect the energy ratio shown in Equation \eqref{equ:dcac_ratio} from trained models published in Torchvision~\cite{marcel2010torchvision}. As Table \ref{tab:dcac} shows, for the convolution kernels in all these networks, we get a ratio greater than one, which means that the energy of DC components is larger than energy of all AC components. Thus, our assumption 
in Section~\ref{sec:err_analysis} empirically holds true in practice.

\section{Conclusions}
\label{sec:conclusion}

In this paper, we have addressed the on-device model training for resource-constrained edge devices. To this end, a new gradient filtering method has been proposed to systematically reduce the computation and memory consumption for the back-propagation algorithm, which is the key bottleneck for efficient model training. 

In Section \ref{sec:method}, a new gradient filtering approach has been proposed to reduce the computation required for propagating gradients through the convolutional layers. The gradient filtering creates an approximate gradient feature map with fewer unique elements and a special structure; this reduces the computation by more than two orders of magnitude. Furthermore, we proved that the error introduced during back-propagation by our gradient filter is bounded so the influence of gradient approximation is limited.

Extensive experiments in Section \ref{sec:exp} have demonstrated the efficiency and wide applicability of our method. Indeed, models can be finetuned with orders of magnitudes fewer computations, while having only a marginal accuracy loss compared to popular baseline methods.

\noindent\textbf{Acknowledgements:} This work was supported in part by the US National Science Foundation (NSF) grant CNS-2007284.

{\small
\bibliographystyle{ieee_fullname}
\bibliography{PaperForReview}
}

\clearpage
\appendix

\noindent In this supplementary material, we present:
\begin{itemize}
    \item \textbf{\ref{sec:supp_ptwd}:} Post-training weight distribution.
    \item \textbf{\ref{sec:supp_gf}:} Detailed derivation for gradient filtering described in Section \ref{sec:method}.
    \item \textbf{\ref{sec:supp_prop}:} Detailed proof for Proposition 1 in Section \ref{sec:err_analysis}.
    \item \textbf{\ref{sec:supp_compr18}:} Visualized computation analysis for ResNet18.
    \item \textbf{\ref{sec:supp_setup}:} Detailed experimental setup for Section \ref{sec:setup}.
    \item \textbf{\ref{sec:supp_seg}:} More experimental results for Semantic Segmentation in Section \ref{sec:exp_seg}.
    \item \textbf{\ref{sec:supp_hyper}:} More experimental results for hyper-parameter exploration on CIFAR datasets in Section \ref{sec:exp_hyper}.
    \item \textbf{\ref{sec:supp_gq}:} Experimental results for combining gradient filtering (our method) with existing INT8 gradient quantization approaches\cite{chen2020statistical, banner2018scalable}.
    \item \textbf{\ref{sec:supp_ondevice}:} More experimental results for on-device performance evaluation in Section \ref{sec:exp_ondevice}.
\end{itemize}

\section{Post-training Weight Distribution}
\label{sec:supp_ptwd}

\begin{figure}[h]
    \centering
    \includegraphics[width=\linewidth]{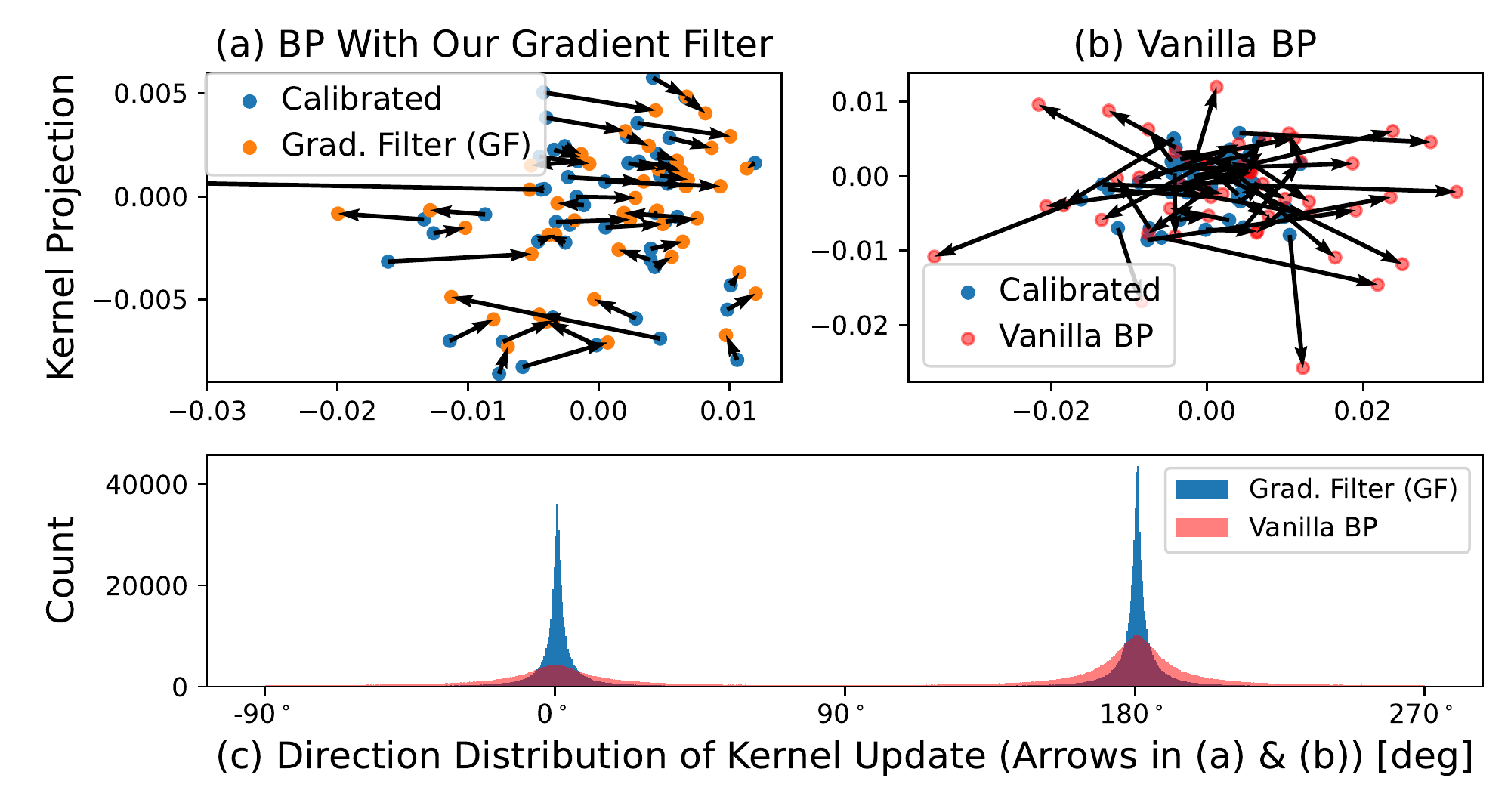}
    \caption{PCA projections of convolution kernels in the bottleneck layer of UperNet. Each point represents a $3\times3$ kernel. (a-b) compare the kernel before training (calibrated) and kernel trained with vanilla BP and our GF. (c) shows the distribution of directions [degree] in which the kernel was updated during training.}
    \label{fig:kern}
\end{figure}

Figure~\ref{fig:kern} shows the PCA projections of convolution kernels in the bottleneck layer of an UperNet.
Since our gradient filter (GF) only keeps the low-frequency part of the gradient signal (see Equation~(\ref{equ:gf_freq})), after applying the gradient filter, only the low-frequency part of the model is updated. As a result, as shown in Figure~\ref{fig:kern}~(a) and (c), using the gradient filter limits the weights update to horizontal directions ($0^\circ$ and $180^\circ$), as opposed to using vanilla back propagation (BP) where all directions are involved (Figure~\ref{fig:kern}~(b) and (c)).

\section{Gradient Filtering Derivation}
\label{sec:supp_gf}
In this section, we present the complete derivations for Equation~(\ref{equ:grad_x}) and Equation~(\ref{equ:grad_w}) in Section \ref{sec:method}, namely the back propagation with gradient filtering. For convenience, Table~\ref{tab:supp_symbols} (reproduced from Table~\ref{tab:symbols} in paper) lists commonly used symbols.
\begin{table}[b]
    \centering
    \begin{tabular}{c|c}
        \toprule
            $C_x$ & Number of channels of $x$ \\
            \hline
            $W_x, H_x$ & Width and height of $x$ \\
            \hline
            $\theta$ & Convolution kernel \\
            \hline
            $\theta'$ & Rotated $\theta$, \textit{i.e.}, $\theta'=\text{rot180}(\theta)$ \\
            \hline
            $r$ & Patch size ($r \times r$) \\
            \hline
            $g_x, g_y, g_\theta$ & Gradients w.r.t. $x, y, \theta$\\
            \hline
            $\tilde g_y$ & Approximated gradient $g_y$ \\
            \hline
            \multirow{2}{*}{$\tilde x, \tilde \theta'$} & Sum of $x$ and $\theta'$ over  \\
            ~ & spatial dimensions (height and width) \\
            \hline
            \multirow{2}{*}{$x[n, c_i, h, w]$} & Element for feature map $x$ \\
            ~ & at batch $n$, channel $c_i$, pixel $(h, w)$\\
            \hline
            \multirow{3}{*}{$\theta[c_o, c_i, u, v]$} & Element for convolution kernel $\theta$ \\
            ~ & at output channel $c_o$, input channel $c_i$,\\
            ~ & position $(u, v)$\\
        \bottomrule
    \end{tabular}
    \caption{Table of symbols we use.}
    \label{tab:supp_symbols}
\end{table}

\subsection{Gradient Filtering}
\label{ssec:grad_filt}
We have:
\begin{equation}
    \tilde g_y[n, c_o, h, w] = \frac{1}{r^2}\sum_{h=\lfloor i/r\rfloor r}^{\lceil i/r\rceil r}\sum_{w=\lfloor j/r\rfloor r}^{\lceil j/r\rceil r} g_y[n, c_o, i, j]
\end{equation}
Thus, for any entry in the approximated gradient $\tilde g_y$, the value equals to the average of all neighboring elements within the same $r\times r$ patch, as shown in Figure \ref{fig:method} in the main manuscript.
For the approximated gradient $\tilde g_y$ with batch size $n$, channel $c$, resolution $(H_y, W_y)$, there will be $(n\times c \times \lceil\frac{H_y}{r}\rceil \times \lceil\frac{W_y}{r}\rceil)$ unique numbers in $\tilde g_y$.
To simplify the following derivations, we rewrite the approximated gradient $\tilde g_y$ as follows:
\begin{equation}
    \tilde g_y^p[n, c_o, h_p, w_p, i, j] = \tilde g_y[n, c_o, h_p * r + i, w_p * r + j]
\end{equation}
where $(h_p, w_p)$ is the position of the patch and $(i, j)$ is the offset within the patch. Since every element in the same patch has the exact same value, we denote this unique value with $\tilde g_y^u$, \textit{i.e.},
\begin{equation}
    \tilde g_y^u[n, c_o, h_p, w_p] = \tilde g_y^p[n, c_o, h_p, w_p, i, j], \forall 0 \le i, j < r
\end{equation}

\subsection{Approximation for Rotated Convolution Kernel $\theta'$}
\begin{equation}
\begin{aligned}
    \tilde \theta'[c_o, c_i] &= \sum_{u, v}\theta'[c_o, c_i, u, v] \\
    &= \sum_{u, v}\text{rot180}(\theta)[c_o, c_i, u, v] \\
    &= \sum_{u, v}\theta[c_o, c_i, u, v]
\end{aligned}
\end{equation}

\subsection{Approximation for Input Feature $x$}
\begin{equation}
    \tilde x[n, c_i, h, w] = \sum_{h=\lfloor i/r\rfloor r}^{\lceil i/r\rceil r}\sum_{w=\lfloor j/r\rfloor r}^{\lceil j/r\rceil r} x[n, c_i, i, j]
\end{equation}
Thus for every entry in approximated feature map $\tilde x$, the value equal to the sum of all neighboring elements within the same $r\times r$ patch. Following the definition of the gradient filter in Section \ref{ssec:grad_filt}, we use the following symbols to simplify the derivation:
\begin{equation}
    \tilde x^p[n, c_i, h_p, w_p, i, j] = \tilde x[n, c_i, h_p * r + i, w_p * r + j]
\end{equation}
and
\begin{equation}
    \tilde x^u[n, c_i, h_p, w_p] = \tilde x^p[n, c_i, h_p, w_p, i, j], \forall 0\le i,j < r
\end{equation}

\subsection{Boundary Elements}
As mentioned in Section \ref{sec:method}, given the structure created by the gradient filters, the gradient propagation in a convolution layer can be simplified to weights summation and multiplication with few unique gradient values. This is true for \textit{all elements} far away from the patch boundary because for these elements, the rotated kernel $\theta'$ only covers the elements from the same patch, which have the same value, thus the computation can be saved. However, for the elements close to the boundary, this is not true, since when convolving with boundary gradient elements, the kernel may cover multiple patches with multiple unique values instead of just one. To eliminate the extra computation introduced by the boundary elements, we pad each patch sufficiently such that every element is far away from boundary:
\begin{equation}
    \tilde g_y^p[n, c_i, h_p, w_p, i, j] = \tilde g_y^u[n, c_i, h_p, w_p], \forall i, j \in \mathbb Z
\end{equation}
For example, with the patch size $4\times 4$, the element at the spatial position $(3, 3)$ is on the boundary, so when we calculate $\tilde g_x[n, c_i, 3, 3]$ by convolving the rotated kernel $\theta'$ with the approximated gradient $\tilde g_y$:
\begin{equation}
    \tilde g_x[n, c_i, 3, 3] = \sum_{i,j}\theta'[c_o, c_i, i, j]\tilde g_y[n, c_o, 3+i, 3+j]
    \label{equ:gx_exp}
\end{equation}
values of $\tilde g_y$ are from multiple patches and have different values (\textit{e.g.}, $\tilde g_y[n, c_o, 3, 3]$ is from patch $(0, 0)$ while $\tilde g_y[n, c_o, 4, 4]$ is from patch $(1, 1)$; they have different values). In our method, we simplify the Equation~\eqref{equ:gx_exp} by rewriting it in the following way:
\begin{align}
&\begin{aligned}
    &\tilde g_x[n, c_i, 3, 3]\\
    &\approx \sum_{i,j=-1}^{1}\theta'[c_o, c_i, i, j]\tilde g_y^p[n, c_o, \lfloor\frac{3}{4}\rfloor, \lfloor\frac{3}{4}\rfloor, 3+i, 3+j]
\end{aligned} \label{equ:gx_exp_approx}\\
    &= \sum_{i,j=-1}^{1}\theta'[c_o, c_i, i, j]\tilde g_y^u[n, c_o, \lfloor\frac{3}{4}\rfloor, \lfloor\frac{3}{4}\rfloor] \\
    &= \sum_{i,j=-1}^{1}\theta'[c_o, c_i, i, j]\tilde g_y^u[n, c_o, 0, 0]
\end{align}
where Equation~\eqref{equ:gx_exp_approx} is derived from Equation~\eqref{equ:gx_exp} by considering that patch $(0, 0)$ is sufficiently padded so that for elements with all offsets $(3+i, 3+j)$, they have the same value, which is the unique value $g_y^u[n, c_o, 0, 0]$.

For approximated input feature map $\tilde x$, we apply the same approximation for the boundary elements.

\subsection{Gradient w.r.t. Input (Equation (\ref{equ:grad_x}) in Section \ref{sec:bpgf})}
\begin{align}
    &\tilde g_x[n, c_i, h, w] \\
    &=\sum_{c_o, u, v}\theta[c_o, c_i, -u, -v]\tilde g_y[n, c_o, h+u, w+v] \label{equ:standard_gx}\\
    &\begin{aligned} \approx&\sum_{c_o, u, v}\theta[c_o, c_i, -u, -v]\cdot\\ &\tilde g_y^p[n, c_o, \lfloor\frac{h}{r}\rfloor, \lfloor\frac{w}{r}\rfloor, (h \text{ mod } r)+u, (w \text{ mod } r) + v]
    \end{aligned} \\
    &=\sum_{c_o, u, v}\theta[c_o, c_i, -u, -v]\tilde g_y^u[n, c_o, \lfloor \frac{h}{r} \rfloor, \lfloor\frac{w}{r}\rfloor] \label{equ:approx_gx}\\
    &=\sum_{c_o}\tilde g_y^u[n, c_o, \lfloor \frac{h}{r} \rfloor, \lfloor\frac{w}{r}\rfloor]\sum_{u, v}\theta[c_o, c_i, -u, -v] \\
    &=\sum_{c_o}\tilde g_y^u[n, c_o, \lfloor \frac{h}{r} \rfloor, \lfloor\frac{w}{r}\rfloor]\tilde \theta'[c_o, c_i] \label{equ:gx_final}
\end{align}
By expanding $\tilde g_y^u$ to $\tilde g_y$, we have:
\begin{equation}
    \tilde g_x[n, c_i, h, w] = \sum_{c_o} \tilde g_y[n, c_o, h, w] \odot \tilde \theta'[c_o, c_i] \label{equ:gx_point}
\end{equation}
which is the Equation (\ref{equ:grad_x}) in Section \ref{sec:method} in the paper.

From Equation~\eqref{equ:standard_gx} to Equation~\eqref{equ:approx_gx}, we consider that the patch in the approximated gradient $\tilde g_y$ is padded sufficiently so they have the same value for all possible offsets $((h \text{ mod } r)+u, (w \text{ mod } r) + v)$. If there is only one input channel and output channel for the convolutional layer as the Figure \ref{fig:method} in the paper shows, then Equation~\eqref{equ:gx_final} become an element-wise multiplication, which is Equation~\eqref{equ:gx_point} (also the Equation~(\ref{equ:grad_x}) in the Section \ref{sec:bpgf}).

\subsection{Gradient w.r.t. Convolution Kernel (Equation (\ref{equ:grad_w}) in the Section \ref{sec:bpgf})}
\begin{align}
    &\tilde g_\theta[c_o, c_i, u, v]\\ 
    &= \sum_{n, h, w} x[n, c_i, h+u, w+v] \tilde g_y[n, c_o, h, w] \label{equ:standard_gw}\\
    &\begin{aligned}\approx \sum_{n, h, w} &\tilde x^p[n, c_i, \lfloor\frac{h}{r}\rfloor, \lfloor\frac{w}{r}\rfloor, (h\text{ mod } r) + u, (w\text{ mod } r) + v] \cdot \\
    &\tilde g_y^u[n, c_o, \lfloor\frac{h}{r}\rfloor, \lfloor\frac{w}{r}\rfloor]
    \end{aligned}\\
    &= \sum_{n, h, w} \tilde x^u[n, c_i, \lfloor\frac{h}{r}\rfloor, \lfloor\frac{w}{r}\rfloor]\tilde g_y^u[n, c_o, \lfloor\frac{h}{r}\rfloor, \lfloor\frac{w}{r}\rfloor] \label{equ:approx_gw}\\
    &= \sum_{n, h, w} \tilde x^u[n, c_i,  \lfloor\frac{h}{r}\rfloor, \lfloor\frac{w}{r}\rfloor]\tilde g_y^u[n, c_o,  \lfloor\frac{h}{r}\rfloor, \lfloor\frac{w}{r}\rfloor] \label{equ:approx_gw_final}
\end{align}
By expanding $\tilde x^u$ and $\tilde g_y^u$ to $\tilde x$ and $\tilde g_y$, respectively, we have:
\begin{equation}
    \tilde g_\theta [c_o, c_i, u, v] = \sum_{n, i, j} \tilde x[n, c_i, i, j]\tilde g_y[n, c_o, i, j]
\end{equation}
which is precisely Equation (\ref{equ:grad_w}) in Section \ref{sec:method}.

From Equation \eqref{equ:standard_gw} to Equation \eqref{equ:approx_gw}, we consider that the patch in the approximated input feature map $\tilde x$ is padded sufficiently thus they have the same value for all possible offsets $((h\text{ mod } r) + u, (w\text{ mod } r) + v)$. 
For every given input/output channel pair $(c_o, c_i)$, Equation~\eqref{equ:approx_gw_final} represents the Frobenius inner product between $\tilde x^u$ and $\tilde g_y^u$.

\section{Detailed Proof for Proposition 1}
\label{sec:supp_prop}
In this section, we provide more details to the proof in Section \ref{sec:err_analysis}.
We use $G_x, G_y$ and $\Theta$ to denote the gradients $g_x, g_y$ and the convolution kernel $\theta$ in the \textit{frequency domain}, respectively. $G_x[u,v]$ is the spectrum value at frequency $(u,v)$ and $\delta$ is the 2D discrete Dirichlet function. Without losing generality and to simplify the proof, we consider the batch size is 1, the number of input/output channels is 1, namely $C_x=C_y=1$, and there is only one patch in $\tilde g_y$.

The gradient returned by the gradient filtering can be written as:
\begin{equation}
    \tilde g_y = \frac{1}{r^2} 1_{r\times r} \circledast g_y
    \label{equ:supp_gf_spatial}
\end{equation}
where $\circledast$ denotes convolution. By applying the discrete Fourier transformation, Equation~\eqref{equ:supp_gf_spatial} can be rewritten in the frequency domain as:
\begin{equation}
\tilde G_y[u,v] = \frac{1}{r^2}\delta[u, v]G_y[u, v]
\label{equ:supp_gf_freq}
\end{equation}
$\tilde g_y$ is the approximation for $g_y$(so the ground truth for $\tilde g_y$ is $g_y$), and the SNR of $\tilde g_y$ equals to:
\begin{equation}
\begin{aligned}
\text{SNR}_{\tilde g_y} &= (\frac{\sum_{(u,v)}(G_y[u,v], - \tilde G_y[u,v])^2}{\sum_{(u,v)} G_y^2[u,v]})^{-1} \\
&= (\frac{\sum_{(u,v)}(G_y[u,v] - \frac{1}{r^2}\delta[u,v]G_y[u,v])^2}{\sum_{(u,v)} G_y^2[u,v]})^{-1} \\
\end{aligned}
\label{equ:basic_snr_gy}
\end{equation}
where the numerator can be written as:
\begin{equation}
    \begin{aligned}
        &\sum_{(u,v)} (G_y[u,v] - \frac{1}{r^2}\delta[u,v]G_y[u,v])^2\\
        & \begin{aligned}
            =&\sum_{(u,v)\ne(0,0)} (G_y[u,v] - \frac{1}{r^2}\delta[u,v]G_y[u,v])^2 \\
            &+ (G_y[0,0] - \frac{1}{r^2}\delta[0,0]G_y[0,0])^2
        \end{aligned}
    \end{aligned}
    \label{equ:numerator_gy_init}
\end{equation}
Because $\delta[u,v]=\begin{cases}1 &(u,v) = (0,0) \\ 0 &(u,v)\ne (0,0)\end{cases}$, Equation~\eqref{equ:numerator_gy_init} can be written as:
\begin{equation}
\begin{aligned}
    &\sum_{(u,v)\ne(0,0)}G_y^2[u,v] + \frac{(r^2-1)^2}{r^4}G_y^2[0,0] \\
    &\begin{aligned}=&\sum_{(u,v)\ne (0,0)}G_y^2[u,v] + G_y^2[0,0] - G_y^2[0,0]\\ &+ \frac{(r^2-1)^2}{r^4}G_y^2[0,0] \end{aligned}\\
    &= \sum_{(u,v)}G_y^2[u,v] - \frac{2r^2-1}{r^4} G_y^2[0,0]
\end{aligned}
\label{equ:numerator_gy}
\end{equation}
By substituting the numerator in Equation~\eqref{equ:basic_snr_gy} with Equation~\eqref{equ:numerator_gy}, we have:
\begin{equation}
\begin{aligned}
\text{SNR}_{\tilde g_y} &= (\frac{\sum_{(u,v)}G_y^2[u,v] - \frac{2r^2-1}{r^4} G_y^2[0,0]}{\sum_{(u,v)} G_y^2[u,v]})^{-1} \\
&= (1 - \frac{2r^2-1}{r^4}\frac{G_y^2[0,0]}{\sum_{(u,v)}G^2_y[u,v]})^{-1} \\
&= (1 - \frac{2r^2-1}{r^4}\frac{\text{Energy of DC Component in } G_y}{\text{Total Energy\footnotemark in } G_y})^{-1}
\end{aligned}
\label{equ:supp_snr_g3}
\end{equation}
\footnotetext{As reminder, the total energy of a signal is the sum of energy in DC component and energy in AC components.}
For the convolution layer, the gradient w.r.t. approximated variable $\tilde x$ in the frequency domain is:
\begin{equation}
\begin{aligned}
	\tilde G_{x}[u,v] &= \Theta[-u,-v]\tilde G_{y}[u,v]\\ 
	&=\frac{1}{r^2}\Theta[-u,-v]\delta[u,v]G_y[u,v]
\end{aligned}
\end{equation}
and its ground truth is:
\begin{equation}
G_{x}[u,v] = \Theta[-u,-v]G_y[u,v]
\end{equation}
Similar to Equation~\eqref{equ:supp_snr_g3}, the SNR of $g_{\tilde x}$ is:
\begin{equation}
\begin{aligned}
\text{SNR}_{\tilde g_x} 
&= (1 - \frac{2r^2-1}{r^4}\frac{\Theta^2[0,0]G_y^2[0,0]}{\sum_{(u,v)}\Theta^2[u,v]G_y^2[u,v]})^{-1} \\
&= (1 - \frac{2r^2-1}{r^4}\frac{G_x^2[0,0]}{\sum_{(u,v)}G_x^2[u,v]})^{-1} \\
&= (1 - \frac{2r^2-1}{r^4}\frac{\text{Energy of DC Component in } G_x}{\text{Total Energy\footnotemark in } G_x})^{-1}  
\end{aligned}
\label{equ:supp_snr_approx_gx}
\end{equation}
Equation~\eqref{equ:supp_snr_approx_gx} can be rewritten as:
\begin{equation}\label{equ:supp_raw_prop}
    \begin{aligned}
        \frac{r^4(1-\text{SNR}^{-1}_{\tilde g_x})}{2r^2-1} &=\frac{(\Theta[0,0]G_y[0,0])^2}{\sum_{(u,v)} (\Theta[-u,-v]G_y[u,v])^2}\\
        &= \frac{G_y^2[0,0]}{\sum_{(u,v)} (\frac{\Theta[-u,-v]}{\Theta[0, 0]}G_y[u,v])^2}
    \end{aligned}
\end{equation}
Besides, the proposition's assumption (the DC component dominates the frequency spectrum of $\Theta$) can be written as:
\begin{equation}
    \frac{\Theta^2[0,0]}{\text{max}_{(u,v)\ne(0,0)} \Theta^2[u,v]} \ge 1
\end{equation}
which is:
\begin{equation}\label{equ:supp_dcac_ratio}
    \forall (u, v), \frac{\Theta^2[-u,-v]}{\Theta^2[0,0]} \le 1
\end{equation}
thus, by combining Equation~\eqref{equ:supp_raw_prop} and Equation~\eqref{equ:supp_dcac_ratio}, we have:
\begin{equation}
\begin{aligned}
    \frac{r^4(1-\text{SNR}^{-1}_{\tilde g_x})}{2r^2-1} &= \frac{G_y^2[0,0]}{\sum_{(u,v)} (\frac{\Theta[-u,-v]}{\Theta[0, 0]}G_y[u,v])^2} \\
    &\ge \frac{G_y^2[0,0]}{\sum_{(u,v)} (G_y[u,v])^2} \\ 
    &= \frac{r^4(1-\text{SNR}^{-1}_{\tilde g_y})}{2r^2-1} \\
\end{aligned}
\end{equation}
which means that: $\text{SNR}_{\tilde g_x} \ge \text{SNR}_{\tilde g_y}$. This completes our proof for error analysis.$\blacksquare$

In conclusion, as the gradient propagates, the noise introduced by the gradient filter becomes weaker and weaker compared to the real gradient signal. This property ensures that the error in gradient has only a limited influence on the quality of BP. 

This proof can be extended to the more general case where batch size and the number of channels are greater than 1 by introducing more dimensions (\textit{i.e.}, batch dimension, channel dimension) into all equations listed above.

\section{Computation Analysis for ResNet18}
\label{sec:supp_compr18}

\begin{figure}
    \centering
    \begin{subfigure}[b]{\linewidth}
         \centering
         \includegraphics[width=0.8\linewidth]{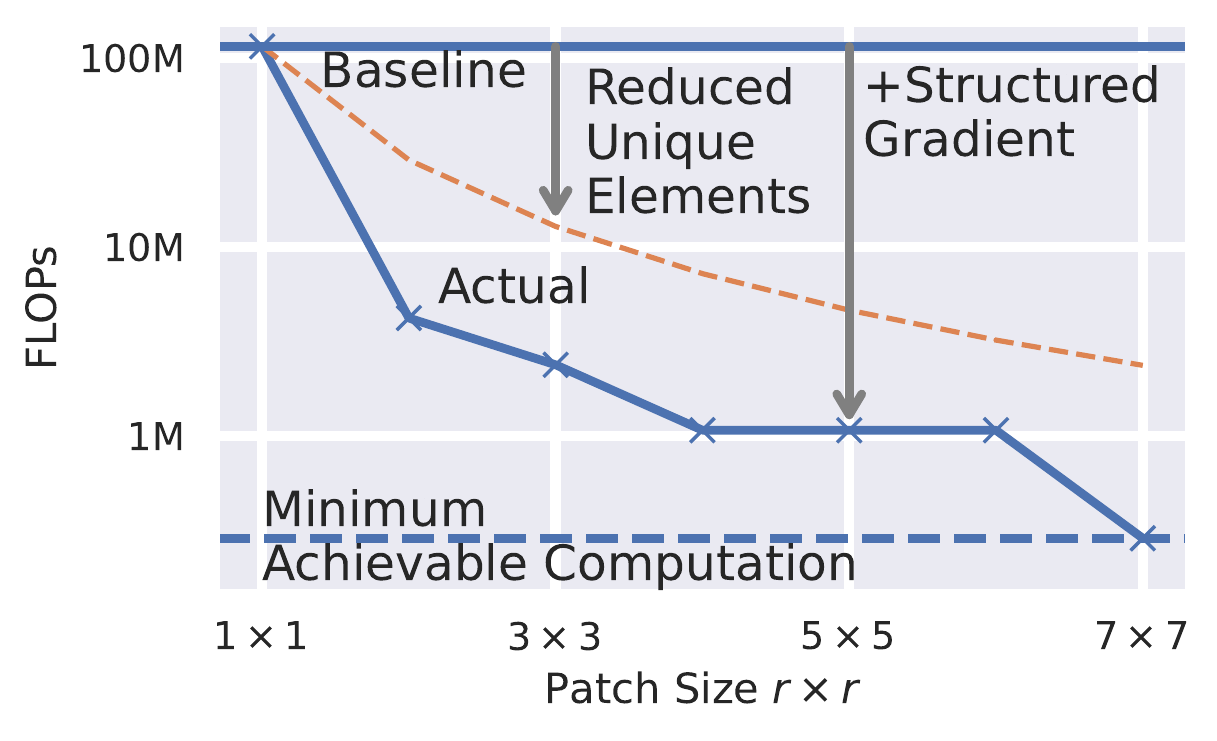}
         \caption{Last convolutional layer in block 4 of ResNet18 with 512 input/output channels; the resolution of input feature map is $7\times 7$.}
         \label{fig:comp_l4}
     \end{subfigure}\\
    \begin{subfigure}[b]{\linewidth}
         \centering
         \includegraphics[width=0.8\linewidth]{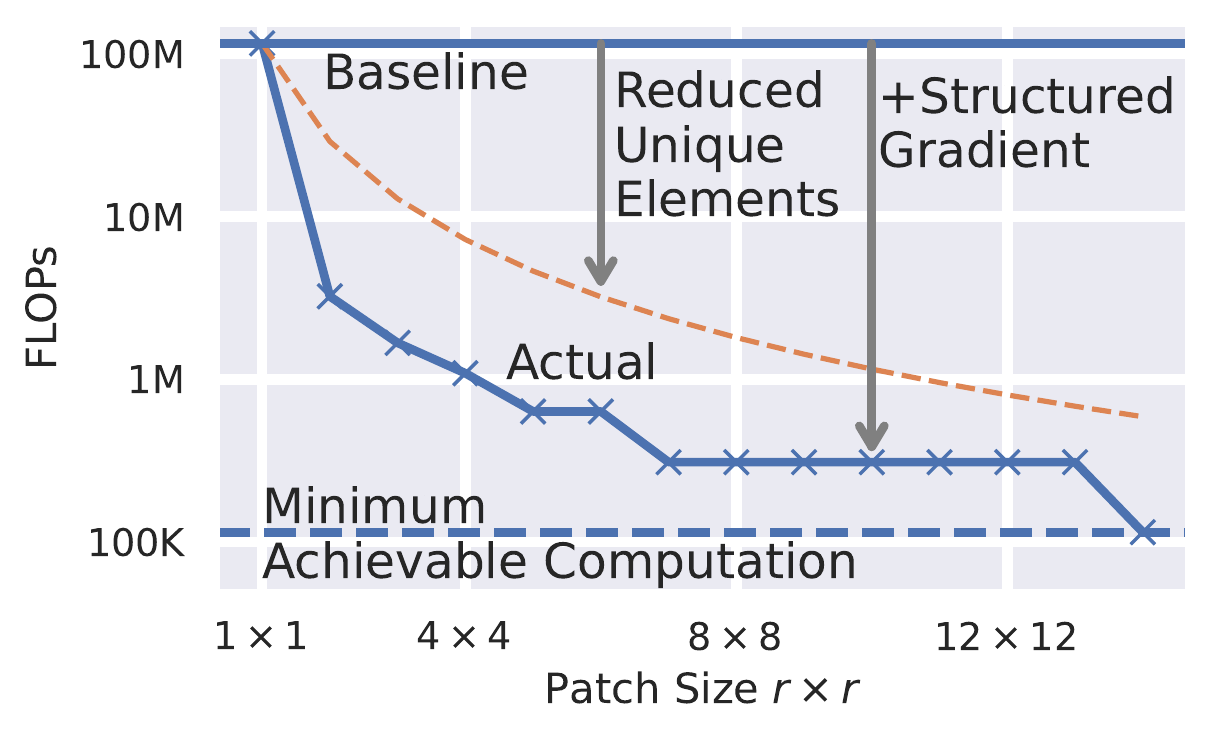}
         \caption{Last convolutional layer in block 3 of ResNet18 with 256 input/output channels; the resolution of input feature map is $14\times 14$.}
         \label{fig:comp_l3}
     \end{subfigure}
    \begin{subfigure}[b]{\linewidth}
         \centering
         \includegraphics[width=0.8\linewidth]{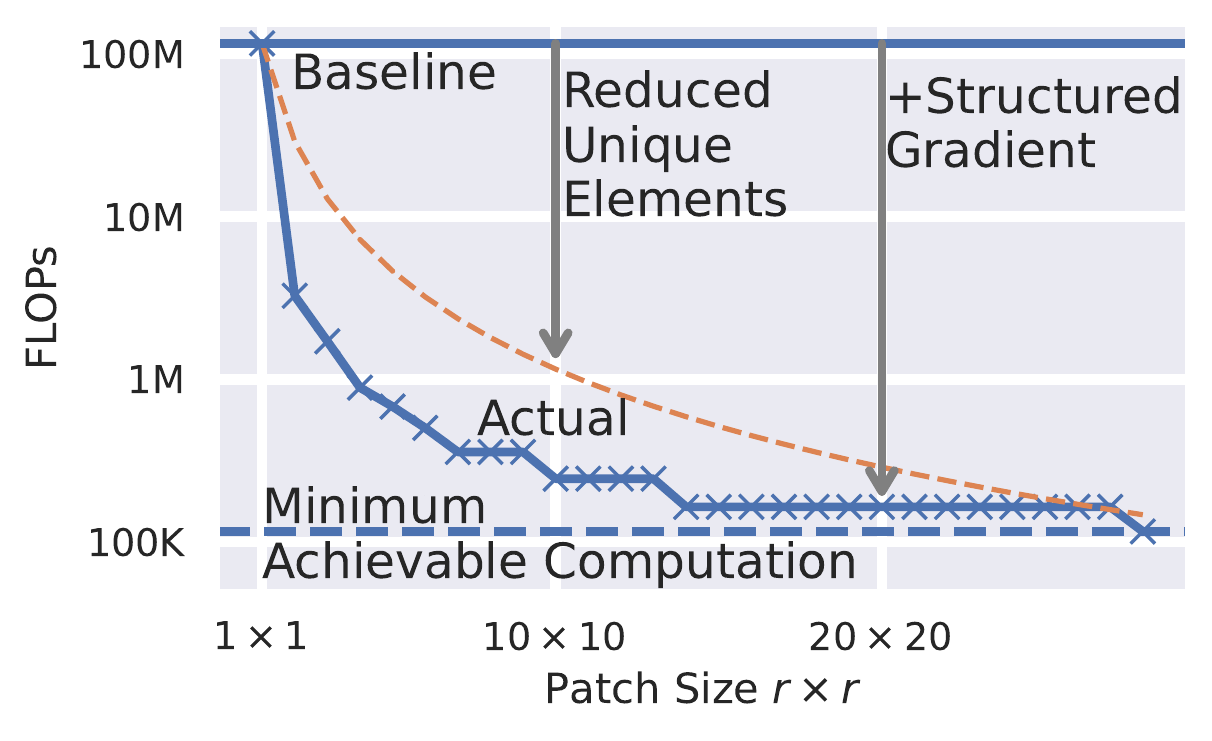}
         \caption{Last convolutional layer in block 2 of ResNet18 with 128 input/output channels; the resolution of input feature map is $28\times 28$.}
         \label{fig:comp_l2}
     \end{subfigure}
    \caption{Computation analysis for three convolution layers in of ResNet18 model. Since convolutional layers in every block of ResNet18 is similar, we use the last convolutional layer as the representative of all convolutional layers in the block. Minimum achievable computation is presented in Equation~(\ref{equ:lb_comp}) in the paper. By reducing the number of unique elements, computations required by our approach drop to about $1/r^2$ compared with the standard BP method. By combining it (“+” in the figure) with structured gradient map, computations required by our approach drop further.}
    \label{fig:comp}
\end{figure}

In this section, we provide one more example for computation analysis in Section \ref{sec:comp_analysis}. Figure~\ref{fig:comp} shows the computation required by the convolution layers from ResNet18 with different patch sizes for gradient filtering. With reduced unique elements, our approach reduces the number of computations to $1/r^2$ of standard BP method; with structured gradient, our approach further reduces the number of computations to about $1/(r^2H_\theta W_\theta)$ of standard BP method.

\section{Detailed Experimental Setup}
\label{sec:supp_setup}
In this section, we extend the experimental setup in Section \ref{sec:setup}.
\subsection{ImageNet Classification}
\subsubsection{Environment}
ImageNet related experiments are conducted on IBM Power System AC922, which is equipped with a 40-core IBM Power 9 CPU, 256 GB DRAM and 4 NVIDIA Tesla V100 16GB GPUs. We use PyTorch 1.9.0 compiled with CUDA 10.1 as the deep learning framework.
\subsubsection{Dataset Split}
\begin{figure}[h]
    \centering
    \includegraphics[width=\linewidth]{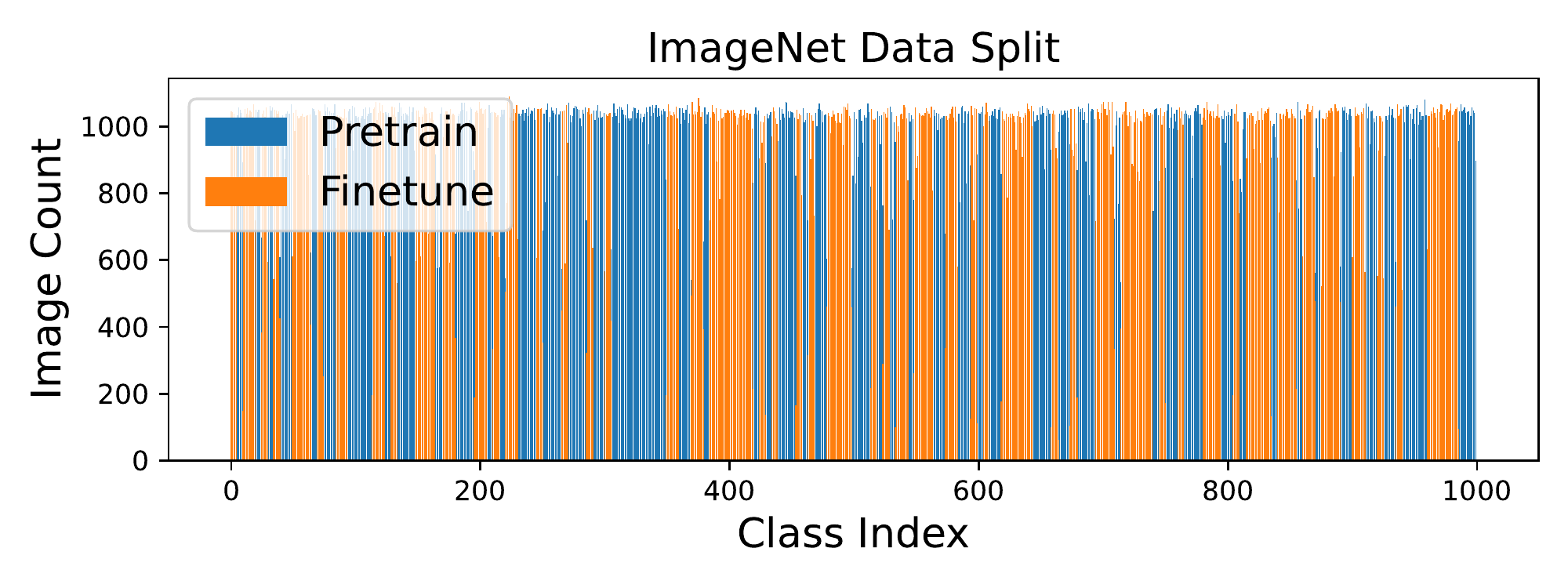}
    \caption{Label distribution for pretraining and finetuning datasets. Pretraining and finetuning partitions are split from ImageNet dataset.}
    \label{fig:imagenet_dist}
\end{figure}

We split the dataset into two non-i.i.d. partitions following the FedAvg method~\cite{mcmahan2017communication}. The label distribution is shown in Figure~\ref{fig:imagenet_dist}. Among all 1000 classes for the ImageNet, pretrain and finetune partitions overlap on only 99 classes, which suggests that our method can efficiently adapt the CNN model to data collected from new environments. For each partition, we randomly select 80\% data as training data and 20\% as validation data.

\subsubsection{Pretraining}
We pretrain ResNet 18, ResNet 34, MobileNet-V2 and MCUNet with the same configuration. We use SGD optimizer. The learning rate of the optimizer starts at 0.05 and decays according to cosine annealing method \cite{loshchilov2017sgdr} during training. Additionally, weight decay is set to $1\times 10^{-4}$ and momentum is set to 0.9.
We set batch size to 64. We randomly resize, randomly flip and normalize the image for data augmentation. We use cross entropy as loss function. Models are trained for 200 epochs and the model with the highest validation accuracy is kept for finetuning. Table \ref{tab:imagenet_pretrain} shows the pretrain accuracy.
\begin{table}[]
\begin{tabular}{cc|cc}
\toprule
Model     & Accuracy & Model        & Accuracy \\ \hline
ResNet-18 & 73.5\%   & MobileNet-V2 & 74.3\%   \\
ResNet-34 & 76.4\%     & MCUNet       & 71.4\%   \\
\bottomrule
\end{tabular}
\caption{Model pretraining accuracy on ImageNet.}
\label{tab:imagenet_pretrain}
\end{table}

\subsubsection{Finetuning}
We adopt the hyper-parameter (\textit{e.g.}, momentum, weight decay, etc.) from pretraining.  Several changes are made: models are finetuned for 90 epochs instead of 200; we apply L2 gradient clipping with threshold 2.0; linear learning rate warm-up for 4 epochs is introduced at the beginning of finetuning, \textit{i.e.}, for the first 4 epochs, the learning rate grows linearly up to 0.05, then the learning rate decays according to cosine annealing method in the following epochs. Of note, to ensure a fair comparison, we use the same hyper-parameters for all experiments, regardless of model type and training strategy.
\subsection{CIFAR Classification}
\subsubsection{Environment} CIFAR related experiments are conducted on a GPU workstation with a 64-core AMD Ryzen Threadripper PRO 3995WX CPU, 512 GB DRAM and 4 NVIDIA RTX A6000 GPUs. We use PyTorch 1.12.0 compiled with CUDA 11.6 as the deep learning framework.
\subsubsection{Dataset Split}
\begin{figure}[h]
    \centering
    \includegraphics[width=\linewidth]{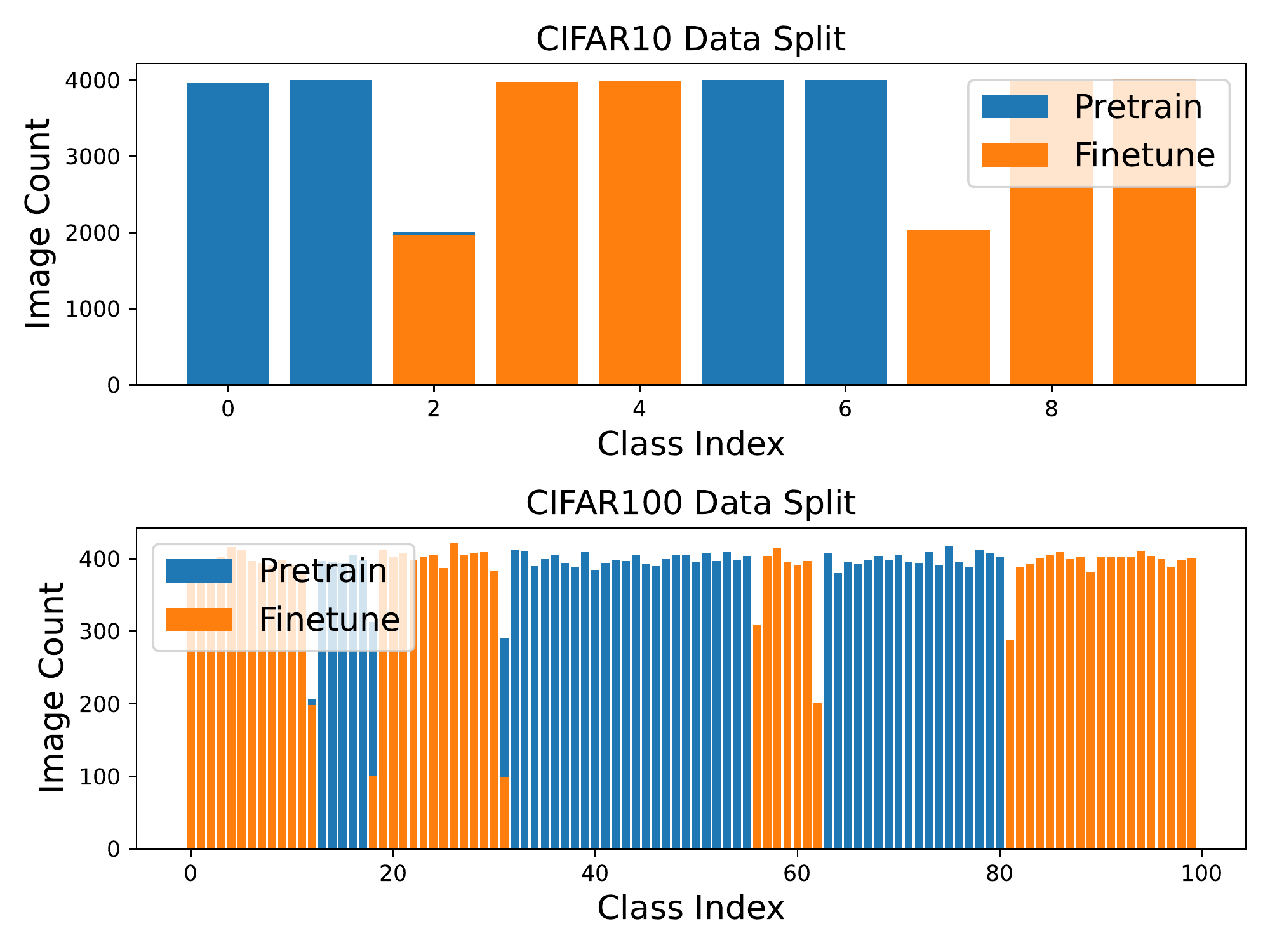}
    \caption{Label distribution for pretraining and finetuning datasets on CIFAR10 and CIFAR100. Pretraining and finetuning partitions are split from CIFAR10/100, respectively.}
    \label{fig:cifar_dist}
\end{figure}

We split the dataset into two non-i.i.d. partitions following FedAvg method. The label distribution is shown in Figure \ref{fig:cifar_dist}. For CIFAR10, pretrain and finetune partitions overlap on 2 classes out of 10 classes in total. For CIFAR100, pretrain and finetune partitions overlap on 6 classes out of 100 classes.

\subsubsection{Pretraining}
We pretrain ResNet18 and ResNet34 with the same configuration. We use the ADAM optimizer with a learning rate of $3\times 10^{-4}$ and weight decay $1\times 10^{-4}$ with no learning rate scheduling method. We use cross entropy as loss function. We set batch size to 128, and normalize the data before feeding it to the model. Models are trained for 30 and 50 epochs for CIFAR10 and CIFAR100, respectively. Then, the model with the highest accuracy is kept for finetuning. Table \ref{tab:cifar_pretrain} shows the pretrain accuracy.

\begin{table}[]
\centering
\begin{tabular}{c|cc}
\toprule
         & ResNet18 & ResNet34 \\ \hline
CIFAR10  & 95.1\%   & 97.6\%   \\
CIFAR100 & 75.5\%   & 83.5\%   \\
\bottomrule
\end{tabular}
\caption{Model pretraining accuracy on CIFAR10/100.}
\label{tab:cifar_pretrain}
\end{table}

\subsubsection{Finetuning}
We adopt the training configuration from PSQ~\cite{chen2020statistical} with some changes. We use cross entropy loss with SGD optimizer for training. The learning rate of the optimizer starts at 0.05 and decays according to cosine annealing method during training. Momentum is set to 0 and weight decay is set to $1\times 10^{-4}$. We apply L2 gradient clipping with a threshold $2.0$. Batch normalization layers are fused with convolution layers before training, which is a common technique for inference acceleration.

\subsection{Semantic Segmentation}
\subsubsection{Environment}
ImageNet related experiments are conducted on IBM Power System AC922, which is equipped with a 40-core IBM Power 9 CPU, 256 GB DRAM and 4 NVIDIA Tesla V100 16GB GPUs. We use PyTorch 1.9.0 compiled with CUDA 10.1 as the deep learning framework. We implement our method based on MMSegmentation 0.27.0.
\subsubsection{Pretraining}
We use models pretrained by MMSegmentation. Considering that the numbers of classes, image statistics, and model hyper-parameters may be different when applying on different datasets, we calibrate the model before finetuning.
We use SGD optimizer. The learning rate of the optimizer starts at 0.01 and decays exponentially during training. Additionally, weight decay is set to $5\times 10^{-4}$ and momentum is set to 0.9. We set batch size to 8. We randomly crop, flip and photo-metric distort and normalize the image for data augmentation. We use cross entropy as loss function.
For DeepLabV3, FCN, PSPNet and UPerNet, we calibrate the classifier (\textit{i.e.}, the last layer) and statistics in batch normalization layers for 1000 steps on the finetuning dataset. For DeepLabV3-MobileNetV2 and PSPNet-MobileNetV2, because the number of channels for convolutional layers in the decoder are different for models applied on different datasets, we calibrate the decoder and statistics in batch normalization layers for 5000 steps on the finetuning dataset. 

\begin{table*}[]
\centering
\resizebox{\textwidth}{!}{%
\begin{tabular}{ccccccccccccccc}
\toprule
\multicolumn{15}{c}{\textbf{Pretrain: ADE20K Finetune: VOC12Aug}}                                                                                                                                                                                                                                                                                                                                                                       \\ \hline
\textbf{UPerNet}               & \multicolumn{1}{c|}{\textbf{\#Layers}} & \textbf{GFLOPs} & \textbf{mIoU}   & \multicolumn{1}{c|}{\textbf{mAcc}}   & \textbf{PSPNet-M}              & \multicolumn{1}{c|}{\textbf{\#Layers}} & \textbf{GFLOPs} & \textbf{mIoU}   & \multicolumn{1}{c|}{\textbf{mAcc}}   & \textbf{DLV3-M}                & \multicolumn{1}{c|}{\textbf{\#Layers}} & \textbf{GFLOPs} & \textbf{mIoU}   & \textbf{mAcc}   \\ \hline
Calibration                    & \multicolumn{1}{c|}{0}                 & 0               & 37.66           & \multicolumn{1}{c|}{50.03}           & Calibration                    & \multicolumn{1}{c|}{0}                 & 0               & 30.93           & \multicolumn{1}{c|}{52.01}           & Calibration                    & \multicolumn{1}{c|}{0}                 & 0               & 35.28           & 56.98           \\ \hline
\multirow{3}{*}{Vanilla BP}    & \multicolumn{1}{c|}{All}               & 541.0           & 67.23{[}0.24{]} & \multicolumn{1}{c|}{79.79{[}0.45{]}} & \multirow{3}{*}{Vanilla BP}    & \multicolumn{1}{c|}{All}               & 42.41           & 53.51{[}0.27{]} & \multicolumn{1}{c|}{67.01{[}0.19{]}} & \multirow{3}{*}{Vanilla BP}    & \multicolumn{1}{c|}{All}               & 54.35           & 60.78{[}0.21{]} & 74.10{[}0.40{]} \\
                               & \multicolumn{1}{c|}{5}                 & 503.9           & 72.01{[}0.09{]} & \multicolumn{1}{c|}{81.97{[}0.30{]}} &                                & \multicolumn{1}{c|}{5}                 & 12.22           & 48.88{[}0.11{]} & \multicolumn{1}{c|}{62.67{[}0.31{]}} &                                & \multicolumn{1}{c|}{5}                 & 14.77           & 51.51{[}0.09{]} & 66.08{[}0.44{]} \\
                               & \multicolumn{1}{c|}{10}                & 507.6           & 72.01{[}0.19{]} & \multicolumn{1}{c|}{81.83{[}0.44{]}} &                                & \multicolumn{1}{c|}{10}                & 22.46           & 53.71{[}0.29{]} & \multicolumn{1}{c|}{67.93{[}0.32{]}} &                                & \multicolumn{1}{c|}{10}                & 33.10           & 57.63{[}0.10{]} & 71.93{[}0.41{]} \\ \hline
\multirow{2}{*}{\textbf{Ours}} & \multicolumn{1}{c|}{5}                 & 1.97            & 71.76{[}0.11{]} & \multicolumn{1}{c|}{81.57{[}0.07{]}} & \multirow{2}{*}{\textbf{Ours}} & \multicolumn{1}{c|}{5}                 & 0.11            & 48.59{[}0.08{]} & \multicolumn{1}{c|}{62.28{[}0.30{]}} & \multirow{2}{*}{\textbf{Ours}} & \multicolumn{1}{c|}{5}                 & 0.26            & 49.40{[}0.00{]} & 64.13{[}0.54{]} \\
                               & \multicolumn{1}{c|}{10}                & 2.22            & 71.78{[}0.23{]} & \multicolumn{1}{c|}{81.55{[}0.38{]}} &                                & \multicolumn{1}{c|}{10}                & 0.76            & 52.77{[}0.37{]} & \multicolumn{1}{c|}{66.82{[}0.47{]}} &                                & \multicolumn{1}{c|}{10}                & 1.40            & 55.14{[}0.15{]} & 69.48{[}0.26{]} \\ \hline
\multicolumn{15}{c}{\textbf{Pretrain: ADE20K Finetune: Cityscapes}}                                                                                                                                                                                                                                                                                                                                                                     \\ \hline
\textbf{UPerNet}               & \multicolumn{1}{c|}{\textbf{\#Layers}} & \textbf{GFLOPs} & \textbf{mIoU}   & \multicolumn{1}{c|}{\textbf{mAcc}}   & \textbf{PSPNet-M}              & \multicolumn{1}{c|}{\textbf{\#Layers}} & \textbf{GFLOPs} & \textbf{mIoU}   & \multicolumn{1}{c|}{\textbf{mAcc}}   & \textbf{DLV3-M}                & \multicolumn{1}{c|}{\textbf{\#Layers}} & \textbf{GFLOPs} & \textbf{mIoU}   & \textbf{mAcc}   \\ \hline
Calibration                    & \multicolumn{1}{c|}{0}                 & 0               & 34.15           & \multicolumn{1}{c|}{42.45}           & Calibration                    & \multicolumn{1}{c|}{0}                 & 0               & 28.83           & \multicolumn{1}{c|}{34.85}           & Calibration                    & \multicolumn{1}{c|}{0}                 & 0               & 41.33           & 48.65           \\ \hline
\multirow{3}{*}{Vanilla BP}    & \multicolumn{1}{c|}{All}               & 1082.1          & 73.02{[}0.14{]} & \multicolumn{1}{c|}{81.01{[}0.20{]}} & \multirow{3}{*}{Vanilla BP}    & \multicolumn{1}{c|}{All}               & 84.82           & 60.21{[}0.40{]} & \multicolumn{1}{c|}{67.72{[}0.68{]}} & \multirow{3}{*}{Vanilla BP}    & \multicolumn{1}{c|}{All}               & 108.7           & 71.12{[}0.14{]} & 79.81{[}0.04{]} \\
                               & \multicolumn{1}{c|}{5}                 & 1007.7          & 62.46{[}0.19{]} & \multicolumn{1}{c|}{72.62{[}0.27{]}} &                                & \multicolumn{1}{c|}{5}                 & 24.43           & 42.09{[}0.43{]} & \multicolumn{1}{c|}{48.70{[}0.49{]}} &                                & \multicolumn{1}{c|}{5}                 & 29.5            & 51.00{[}0.05{]} & 59.20{[}0.03{]} \\
                               & \multicolumn{1}{c|}{10}                & 1015.3          & 64.01{[}0.21{]} & \multicolumn{1}{c|}{73.11{[}0.32{]}} &                                & \multicolumn{1}{c|}{10}                & 44.90           & 54.03{[}0.24{]} & \multicolumn{1}{c|}{61.48{[}0.10{]}} &                                & \multicolumn{1}{c|}{10}                & 66.2            & 61.02{[}0.14{]} & 69.80{[}0.06{]} \\ \hline
\multirow{2}{*}{\textbf{Ours}} & \multicolumn{1}{c|}{5}                 & 3.94            & 60.58{[}0.25{]} & \multicolumn{1}{c|}{70.67{[}0.32{]}} & \multirow{2}{*}{\textbf{Ours}} & \multicolumn{1}{c|}{5}                 & 0.22            & 41.59{[}0.38{]} & \multicolumn{1}{c|}{48.10{[}0.41{]}} & \multirow{2}{*}{\textbf{Ours}} & \multicolumn{1}{c|}{5}                 & 0.50            & 48.83{[}0.07{]} & 56.87{[}0.08{]} \\
                               & \multicolumn{1}{c|}{10}                & 4.43            & 62.14{[}0.24{]} & \multicolumn{1}{c|}{71.41{[}0.27{]}} &                                & \multicolumn{1}{c|}{10}                & 1.51            & 49.10{[}0.49{]} & \multicolumn{1}{c|}{56.93{[}1.43{]}} &                                & \multicolumn{1}{c|}{10}                & 2.74            & 50.22{[}1.01{]} & 59.99{[}0.31{]} \\ \bottomrule
\end{tabular}%
}
\caption{Experimental results for semantic segmentation task for UPerNet, DeepLabV3-MobileNetV2 (DLV3-M) and PSPNet-MobileNetV2 (PSPNet-M). Models are pretrained on ADE20K dataset and finetuned on augmentated Pascal VOC12 dataset and Cityscapes dataset respectively. ``\#Layers'' is short for ``the number of \textit{active} convolutional layers'' that are trained. Strategy ``Calibration'' shows the accuracy when only the classifier and normalization statistics are updated to adapt differences (\textit{e.g.} different number of classes) between pretraining dataset and finetuning dataset.}
\label{tab:seg_supp}
\end{table*}

\subsubsection{Finetuning}
We finetune all models with the same configuration. We use the SGD optimizer. The learning rate of the optimizer starts at 0.01 and decays according to cosine anneling method during training. Additionally, weight decay is set to $5\times 10^{-4}$ and momentum is set to 0.9. We set batch size to 8. We randomly crop, flip and photo-metric distort and normalize the image for data augmentation. We use cross entropy as loss function. Models are finetuned for 20000 steps. Experiments are repeated three times with random seed 233, 234 and 235.

\subsection{On-device Performance Evaluation}
\subsubsection{NVIDIA Jetson Nano}
We use NVIDIA Jetson Nano with quad-core Cortex-A57, 4 GB DRAM, 128-core Maxwell edge GPU for performance evaluation on both edge CPU and edge GPU. We use the aarch64-OS Ubuntu 18.04.6 provided by NVIDIA. During evaluation, the frequencies for CPU and GPU are 1.5 GHz and 921 MHz, respectively. Our code and library MKLDNN (a.k.a. OneDNN) are compiled on Jetson Nano with GCC 7.5.0, while libraries CUDA and CUDNN are compiled by NVIDIA.
For CPU evaluations, our code and baseline are implemented with MKLDNN v2.6.
For GPU evaluations, our code and baseline are implemented with CUDA 10.2 and CUDNN 8.2.1.

Before the evaluation for every test case, we warm up the device by running the test once. Then we repeat the test 10 times and report the average value for latency, energy consumption, etc.

Energy consumption is obtained by reading the embedded power meter in Jetson Nano every 20 ms.
\subsubsection{Raspberry Pi-3b}
We use Raspberry Pi-3b with quad-core Cortex-A53, 1 GB DRAM for performance evaluation on CPU. We use the aarch64-OS Raspberry Pi OS. During evaluation, the frequency for CPU is 1.2 GHz. Our code and library MKLDNN are compiled on Raspberry Pi with GCC 10.2. Our code and baseline are implemented with MKLDNN v2.6.

Before the evaluation for every test case, we warm up the device by running the test once. Then we repeat the test 10 times and report the average value for latency, etc.

\begin{table}[]
\centering
\resizebox{\linewidth}{!}{%
\begin{tabular}{c|cccc}
\toprule
\textbf{No.} & \textbf{\#Input Channel} & \textbf{\#Output Channel} & \textbf{Input Width} & \textbf{Input Height} \\ \hline
0   & 128             & 128              & 120         & 160          \\
1   & 256             & 256              & 60          & 80           \\
2   & 512             & 512              & 30          & 40           \\
3   & 512             & 512              & 14          & 14           \\
4   & 256             & 256              & 14          & 14           \\
5   & 128             & 128              & 28          & 28           \\
6   & 64              & 64               & 56          & 56          \\
\bottomrule
\end{tabular}%
}
\caption{Layer configuration for test cases in Figure \ref{fig:speedup_mem} in Section \ref{sec:exp_ondevice} in the paper.}
\label{tab:eval_cfg}
\end{table}

\begin{table*}[h]
\centering
\resizebox{\textwidth}{!}{%
\begin{tabular}{cccccccc|cccccccc}
\toprule
\multicolumn{8}{c|}{\textbf{CIFAR10}}                                                                                                                                                                                                                                                                           & \multicolumn{8}{c}{\textbf{CIFAR100}}                                                                                                                                                                                                                                                                           \\ \hline
\textbf{ResNet18}                                                                        & \multicolumn{1}{c|}{\#Layers} & ACC{[}\%{]} & \multicolumn{1}{c|}{FLOPs}   &\textbf{ResNet34}                                                                        & \multicolumn{1}{c|}{\#Layers} & ACC{[}\%{]} & FLOPs   & \textbf{ResNet18}                                                                        & \multicolumn{1}{c|}{\#Layers} & ACC{[}\%{]} & \multicolumn{1}{c|}{FLOPs}   & \textbf{ResNet34}                                                                        & \multicolumn{1}{c|}{\#Layers} & ACC{[}\%{]} & FLOPs   \\ \hline
\multirow{4}{*}{\begin{tabular}[c]{@{}c@{}}Vanilla\\ BP\end{tabular}}           & \multicolumn{1}{c|}{1}        & 91.7        & \multicolumn{1}{c|}{128.25M} & \multirow{4}{*}{\begin{tabular}[c]{@{}c@{}}Vanilla\\ BP\end{tabular}}           & \multicolumn{1}{c|}{1}        & 94.2        & 128.25M & \multirow{4}{*}{\begin{tabular}[c]{@{}c@{}}Vanilla\\ BP\end{tabular}}           & \multicolumn{1}{c|}{1}        & 73.8        & \multicolumn{1}{c|}{128.39M} & \multirow{4}{*}{\begin{tabular}[c]{@{}c@{}}Vanilla\\ BP\end{tabular}}           & \multicolumn{1}{c|}{1}        & 76.9        & 128.39M \\
                                                                                & \multicolumn{1}{c|}{2}        & 93.6        & \multicolumn{1}{c|}{487.68M} &                                                                                 & \multicolumn{1}{c|}{2}        & 96.6        & 487.68M &                                                                                 & \multicolumn{1}{c|}{2}        & 77.6        & \multicolumn{1}{c|}{487.82M} &                                                                                 & \multicolumn{1}{c|}{2}        & 82.0          & 487.82M \\
                                                                                & \multicolumn{1}{c|}{3}        & 93.7        & \multicolumn{1}{c|}{847.15M} &                                                                                 & \multicolumn{1}{c|}{3}        & 96.6        & 847.13M &                                                                                 & \multicolumn{1}{c|}{3}        & 77.6        & \multicolumn{1}{c|}{847.29M} &                                                                                 & \multicolumn{1}{c|}{3}        & 82.1        & 847.27M \\
                                                                                & \multicolumn{1}{c|}{4}        & 94.4        & \multicolumn{1}{c|}{1.14G}   &                                                                                 & \multicolumn{1}{c|}{4}        & 96.8        & 1.21G   &                                                                                 & \multicolumn{1}{c|}{4}        & 78.0          & \multicolumn{1}{c|}{1.14G}   &                                                                                 & \multicolumn{1}{c|}{4}        & 83.0          & 1.21G   \\ \hline
\multirow{4}{*}{\begin{tabular}[c]{@{}c@{}}+Gradient\\ Filter\\ R2\end{tabular}} & \multicolumn{1}{c|}{1}        & 91.5        & \multicolumn{1}{c|}{8.18M}   & \multirow{4}{*}{\begin{tabular}[c]{@{}c@{}}+Gradient\\ Filter\\ R2\end{tabular}} & \multicolumn{1}{c|}{1}        & 94.2        & 8.18M   & \multirow{4}{*}{\begin{tabular}[c]{@{}c@{}}+Gradient\\ Filter\\ R2\end{tabular}} & \multicolumn{1}{c|}{1}        & 73.7        & \multicolumn{1}{c|}{8.31M}   & \multirow{4}{*}{\begin{tabular}[c]{@{}c@{}}+Gradient\\ Filter\\ R2\end{tabular}} & \multicolumn{1}{c|}{1}        & 77.0          & 8.31M   \\
                                                                                & \multicolumn{1}{c|}{2}        & 92.7        & \multicolumn{1}{c|}{26.80M}  &                                                                                 & \multicolumn{1}{c|}{2}        & 96.6        & 26.80M  &                                                                                 & \multicolumn{1}{c|}{2}        & 75.6        & \multicolumn{1}{c|}{26.94M}  &                                                                                 & \multicolumn{1}{c|}{2}        & 81.1        & 26.94M  \\
                                                                                & \multicolumn{1}{c|}{3}        & 92.8        & \multicolumn{1}{c|}{45.45M}  &                                                                                 & \multicolumn{1}{c|}{3}        & 96.5        & 45.44M  &                                                                                 & \multicolumn{1}{c|}{3}        & 75.6        & \multicolumn{1}{c|}{45.59M}  &                                                                                 & \multicolumn{1}{c|}{3}        & 81.1        & 45.58M  \\
                                                                                & \multicolumn{1}{c|}{4}        & 93.9        & \multicolumn{1}{c|}{60.01M}  &                                                                                 & \multicolumn{1}{c|}{4}        & 96.6        & 64.07M  &                                                                                 & \multicolumn{1}{c|}{4}        & 76.4        & \multicolumn{1}{c|}{60.15M}  &                                                                                 & \multicolumn{1}{c|}{4}        & 82.0          & 64.21M  \\ \hline
\multirow{4}{*}{\begin{tabular}[c]{@{}c@{}}+Gradient\\ Filter\\ R4\end{tabular}} & \multicolumn{1}{c|}{1}        & 91.4        & \multicolumn{1}{c|}{1.88M}   & \multirow{4}{*}{\begin{tabular}[c]{@{}c@{}}+Gradient\\ Filter\\ R4\end{tabular}} & \multicolumn{1}{c|}{1}        & 94.3        & 1.88M   & \multirow{4}{*}{\begin{tabular}[c]{@{}c@{}}+Gradient\\ Filter\\ R4\end{tabular}} & \multicolumn{1}{c|}{1}        & 73.7        & \multicolumn{1}{c|}{2.02M}   & \multirow{4}{*}{\begin{tabular}[c]{@{}c@{}}+Gradient\\ Filter\\ R4\end{tabular}} & \multicolumn{1}{c|}{1}        & 76.9        & 2.02M   \\
                                                                                & \multicolumn{1}{c|}{2}        & 92.7        & \multicolumn{1}{c|}{7.93M}   &                                                                                 & \multicolumn{1}{c|}{2}        & 96.4        & 7.93M   &                                                                                 & \multicolumn{1}{c|}{2}        & 74.9        & \multicolumn{1}{c|}{8.07M}   &                                                                                 & \multicolumn{1}{c|}{2}        & 80.4        & 8.07M   \\
                                                                                & \multicolumn{1}{c|}{3}        & 92.8        & \multicolumn{1}{c|}{13.99M}  &                                                                                 & \multicolumn{1}{c|}{3}        & 96.4        & 13.98M  &                                                                                 & \multicolumn{1}{c|}{3}        & 74.9        & \multicolumn{1}{c|}{14.12M}  &                                                                                 & \multicolumn{1}{c|}{3}        & 80.4        & 14.12M  \\
                                                                                & \multicolumn{1}{c|}{4}        & 93.3        & \multicolumn{1}{c|}{19.12M}  &                                                                                 & \multicolumn{1}{c|}{4}        & 96.1        & 20.04M  &                                                                                 & \multicolumn{1}{c|}{4}        & 75.2        & \multicolumn{1}{c|}{19.26M}  &                                                                                 & \multicolumn{1}{c|}{4}        & 80.5        & 20.17M  \\ \hline
\multirow{4}{*}{\begin{tabular}[c]{@{}c@{}}+Gradient\\ Filter\\ R7\end{tabular}} & \multicolumn{1}{c|}{1}        & 91.5        & \multicolumn{1}{c|}{303.10K} & \multirow{4}{*}{\begin{tabular}[c]{@{}c@{}}+Gradient\\ Filter\\ R7\end{tabular}} & \multicolumn{1}{c|}{1}        & 94.2        & 303.10K & \multirow{4}{*}{\begin{tabular}[c]{@{}c@{}}+Gradient\\ Filter\\ R7\end{tabular}} & \multicolumn{1}{c|}{1}        & 73.7        & \multicolumn{1}{c|}{441.34K} & \multirow{4}{*}{\begin{tabular}[c]{@{}c@{}}+Gradient\\ Filter\\ R7\end{tabular}} & \multicolumn{1}{c|}{1}        & 76.9        & 441.34K \\
                                                                                & \multicolumn{1}{c|}{2}        & 91.5        & \multicolumn{1}{c|}{3.21M}   &                                                                                 & \multicolumn{1}{c|}{2}        & 95.8        & 3.21M   &                                                                                 & \multicolumn{1}{c|}{2}        & 74.1        & \multicolumn{1}{c|}{3.35M}   &                                                                                 & \multicolumn{1}{c|}{2}        & 80.4        & 3.35M   \\
                                                                                & \multicolumn{1}{c|}{3}        & 91.7        & \multicolumn{1}{c|}{6.12M}   &                                                                                 & \multicolumn{1}{c|}{3}        & 96.0          & 6.12M   &                                                                                 & \multicolumn{1}{c|}{3}        & 74.1        & \multicolumn{1}{c|}{6.26M}   &                                                                                 & \multicolumn{1}{c|}{3}        & 80.3        & 6.26M   \\
                                                                                & \multicolumn{1}{c|}{4}        & 92.6        & \multicolumn{1}{c|}{8.90M}   &                                                                                 & \multicolumn{1}{c|}{4}        & 96.0          & 9.03M   &                                                                                 & \multicolumn{1}{c|}{4}        & 75.4        & \multicolumn{1}{c|}{9.04M}   &                                                                                 & \multicolumn{1}{c|}{4}        & 80.3        & 9.17M  \\
                                                                                \bottomrule
\end{tabular}%
}
\caption{Experimental results on CIFAR10 and CIFAR100 datasets for ResNet18 and ResNet34 with different hyper-parameter selections. ``ACC'' is short for accuracy. ``\#Layers'' is short for ``the number of \textit{active} convolution layers''. For example. \#Layers equals to 2 means that only the last two convolutional layers are trained. ``Gradient Filter R2/4/7'' use proposed gradient filtering method with patch size $2\times 2$, $4\times 4$ and $7\times 7$, respectively.}
\label{tab:cifar_hyper}
\end{table*}

\subsubsection{Desktop}
We use a desktop PC with Intel 11900KF CPU, 32 GB DRAM and RTX 3090 Ti GPU for perforamce evaluation on both desktop CPU and desktop GPU. We use x86\_64-OS Ubuntu 20.04. During evaluation, the frequencies for CPU and GPU are 4.7 GHz and 2.0 GHz respectively. Our code is compiled with GCC 9.4.0. MKLDNN is compiled by Anaconda (tag omp\_h13be974\_0). CUDA and CUDNN are compiled by NVIDIA.
For CPU evaluations, our code and baseline are implemented with MKLDNN v2.6.
For GPU evaluations, our code and baseline are implemented with CUDA 11.7 and CUDNN 8.2.1.

Before the evaluation for every test case, we warm up the device by running the 10 times. Then we repeat the test 200 times and report the average value for latency, etc.

\subsubsection{Test Case Configurations}
Table \ref{tab:eval_cfg} lists the configurations for test cases shown in Figure \ref{fig:speedup_mem} in the paper. In addition to the parameters shown in the table, for all test cases, we set the batch size to 32, kernel size to $3\times 3$, padding and stride to 1.

\section{More Results for Semantic Segmentation}
\label{sec:supp_seg}
In this section, we extend the experimental results shown in Section \ref{sec:exp_seg} (Table \ref{tab:seg}). Table \ref{tab:seg_supp} shows the experimental results for UPerNet, PSPNet-MobileNetV2 (PSPNet-M) and DeepLabV3-MobileNetV2 (DLV3-M) on two pairs of pretraing and finetuning datasets. These results further show the effectiveness of our method on a dense prediction task.

\section{More Results for CIFAR10/100 with Different Hyper-Parameter Selections}
\label{sec:supp_hyper}
In this section, we extend the experimental results shown in Section \ref{sec:exp_hyper} (Figure \ref{fig:hyper_parameter}). Table \ref{tab:cifar_hyper} shows the experimental results for ResNet18 and ResNet34 on CIFAR datasets. For every model, we test our method with different patch sizes for gradient filtering and different numbers of \textit{active} convolutional layers (\#Layers in Table \ref{tab:cifar_hyper}, \textit{e.g.}, if \#Layers equals to 2, the last two convolutional layers are trained while other layers are frozen). These results further support the qualitative findings in Section \ref{sec:exp_hyper}.

\begin{figure*}[tbp]
    \centering
    \includegraphics[width=\linewidth]{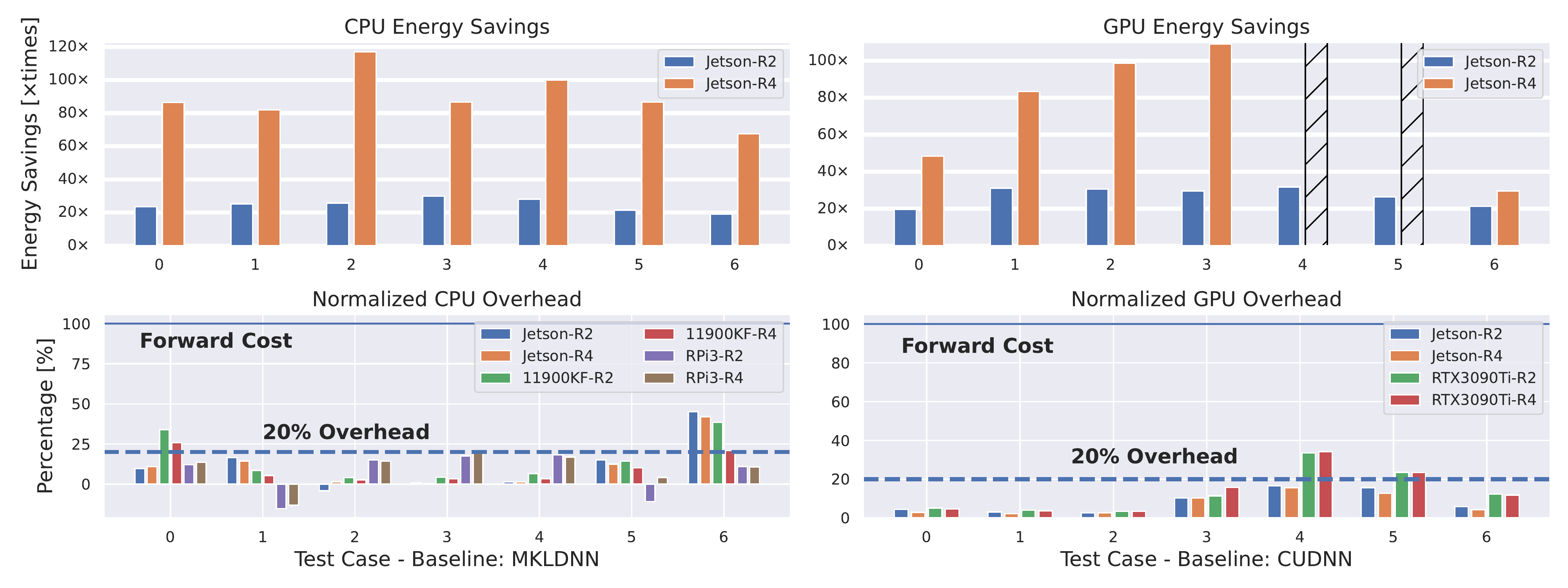}
    \caption{Energy savings and overhead resuls on multiple CPUs and GPUs under different test cases (\textit{i.e.}, different input sizes, number of channels, etc..). For test case 4 and 5 with patch size $4\times 4$ (Jetson-R4) on GPU, the latency of our method is too small to be captured by the power meter with a 20 ms sample rate so the energy savings data is not available. For most test cases with patch size $4\times 4$, our method achieves over $80\times$ energy savings with less than 20\% overhead.}
    \label{fig:overhead_energy}
\end{figure*}

\section{Results for Combining Gradient Filtering with Gradient Quantization}
\label{sec:supp_gq}
In this section, we provide experimental results for combining our method, \textit{i.e.} gradient filtering, with gradient quantization. Table \ref{tab:cifar_gq} shows experimental results for ResNet18 and ResNet32 with gradient quantization methods PTQ~\cite{banner2018scalable} and PSQ~\cite{chen2020statistical} and different hyper-parameters. Both forward propagation and backward propagation are quantized to INT8. These results support the wide applicability of our method.

\section{More Results for On-device Performance Evaluation}
\label{sec:supp_ondevice}

In this section, we extend the experimental results shown in Section \ref{sec:exp_ondevice}.
Figure \ref{fig:overhead_energy} shows the energy savings and overhead of our method. For most test cases with patch $4\times 4$, we achieve over $80\times$ energy savings with less than 20\% overhead on both CPU and GPU. Moreover, for the test case 1 on Raspberry Pi-3b CPU, the forward propagation is even faster when applied our method (which results in negtive overheads). These results further show that our method is practical for the real deployment of both high-performance and IoT applications.

\begin{table*}[]
\centering
\resizebox{\textwidth}{!}{%
\begin{tabular}{cccccccc|cccccccc}
\toprule
\multicolumn{8}{c|}{\textbf{CIFAR10}}                                                                                                                                                                                                                                                                                         & \multicolumn{8}{c}{\textbf{CIFAR100}}                                                                                                                                                                                                                                                                                         \\ \hline
\multicolumn{4}{c|}{\textbf{ResNet18}}                                                                                                                                       & \multicolumn{4}{c|}{\textbf{ResNet34}}                                                                                                                  & \multicolumn{4}{c|}{\textbf{ResNet18}}                                                                                                                                       & \multicolumn{4}{c}{\textbf{ResNet34}}                                                                                                                   \\ \hline
Strategy                                                                               & \multicolumn{1}{c|}{\#Layers} & ACC{[}\%{]} & \multicolumn{1}{c|}{\#OPs}   & Strategy                                                                               & \multicolumn{1}{c|}{\#Layers} & ACC{[}\%{]} & \#OPs   & Strategy                                                                               & \multicolumn{1}{c|}{\#Layers} & ACC{[}\%{]} & \multicolumn{1}{c|}{\#OPs}   & Strategy                                                                               & \multicolumn{1}{c|}{\#Layers} & ACC{[}\%{]} & \#OPs   \\ \hline
\multirow{4}{*}{PTQ}                                                                   & \multicolumn{1}{c|}{1}        & 91.6        & \multicolumn{1}{c|}{128.25M} & \multirow{4}{*}{PTQ}                                                                   & \multicolumn{1}{c|}{1}        & 93.6        & 128.25M & \multirow{4}{*}{PTQ}                                                                   & \multicolumn{1}{c|}{1}        & 74.0        & \multicolumn{1}{c|}{128.39M} & \multirow{4}{*}{PTQ}                                                                   & \multicolumn{1}{c|}{1}        & 76.4        & 128.39M \\
                                                                                       & \multicolumn{1}{c|}{2}        & 93.2        & \multicolumn{1}{c|}{487.68M} &                                                                                        & \multicolumn{1}{c|}{2}        & 96.2        & 487.68M &                                                                                        & \multicolumn{1}{c|}{2}        & 77.8        & \multicolumn{1}{c|}{487.82M} &                                                                                        & \multicolumn{1}{c|}{2}        & 80.3        & 487.82M \\
                                                                                       & \multicolumn{1}{c|}{3}        & 93.5        & \multicolumn{1}{c|}{847.15M} &                                                                                        & \multicolumn{1}{c|}{3}        & 96.2        & 847.13M &                                                                                        & \multicolumn{1}{c|}{3}        & 77.9        & \multicolumn{1}{c|}{847.29M} &                                                                                        & \multicolumn{1}{c|}{3}        & 80.5        & 847.27M \\
                                                                                       & \multicolumn{1}{c|}{4}        & 94.4        & \multicolumn{1}{c|}{1.14G}   &                                                                                        & \multicolumn{1}{c|}{4}        & 96.5        & 1.21G   &                                                                                        & \multicolumn{1}{c|}{4}        & 77.9        & \multicolumn{1}{c|}{1.14G}   &                                                                                        & \multicolumn{1}{c|}{4}        & 82.2        & 1.21G   \\ \hline
\multirow{4}{*}{\begin{tabular}[c]{@{}c@{}}PTQ\\ +Gradient\\ Filter\\ R2\end{tabular}} & \multicolumn{1}{c|}{1}        & 91.4        & \multicolumn{1}{c|}{8.18M}   & \multirow{4}{*}{\begin{tabular}[c]{@{}c@{}}PTQ\\ +Gradient\\ Filter\\ R2\end{tabular}} & \multicolumn{1}{c|}{1}        & 93.5        & 8.18M   & \multirow{4}{*}{\begin{tabular}[c]{@{}c@{}}PTQ\\ +Gradient\\ Filter\\ R2\end{tabular}} & \multicolumn{1}{c|}{1}        & 73.9        & \multicolumn{1}{c|}{8.31M}   & \multirow{4}{*}{\begin{tabular}[c]{@{}c@{}}PTQ\\ +Gradient\\ Filter\\ R2\end{tabular}} & \multicolumn{1}{c|}{1}        & 76.5        & 8.31M   \\
                                                                                       & \multicolumn{1}{c|}{2}        & 92.6        & \multicolumn{1}{c|}{26.80M}  &                                                                                        & \multicolumn{1}{c|}{2}        & 95.9        & 26.80M  &                                                                                        & \multicolumn{1}{c|}{2}        & 75.7        & \multicolumn{1}{c|}{26.94M}  &                                                                                        & \multicolumn{1}{c|}{2}        & 80.0        & 26.94M  \\
                                                                                       & \multicolumn{1}{c|}{3}        & 92.7        & \multicolumn{1}{c|}{45.45M}  &                                                                                        & \multicolumn{1}{c|}{3}        & 96.0        & 45.44M  &                                                                                        & \multicolumn{1}{c|}{3}        & 75.9        & \multicolumn{1}{c|}{45.59M}  &                                                                                        & \multicolumn{1}{c|}{3}        & 80.1        & 45.58M  \\
                                                                                       & \multicolumn{1}{c|}{4}        & 93.7        & \multicolumn{1}{c|}{60.01M}  &                                                                                        & \multicolumn{1}{c|}{4}        & 96.2        & 64.07M  &                                                                                        & \multicolumn{1}{c|}{4}        & 76.3        & \multicolumn{1}{c|}{60.15M}  &                                                                                        & \multicolumn{1}{c|}{4}        & 80.9        & 64.21M  \\ \hline
\multirow{4}{*}{\begin{tabular}[c]{@{}c@{}}PTQ\\ +Gradient\\ Filter\\ R4\end{tabular}} & \multicolumn{1}{c|}{1}        & 91.3        & \multicolumn{1}{c|}{1.88M}   & \multirow{4}{*}{\begin{tabular}[c]{@{}c@{}}PTQ\\ +Gradient\\ Filter\\ R4\end{tabular}} & \multicolumn{1}{c|}{1}        & 93.6        & 1.88M   & \multirow{4}{*}{\begin{tabular}[c]{@{}c@{}}PTQ\\ +Gradient\\ Filter\\ R4\end{tabular}} & \multicolumn{1}{c|}{1}        & 73.7        & \multicolumn{1}{c|}{2.02M}   & \multirow{4}{*}{\begin{tabular}[c]{@{}c@{}}PTQ\\ +Gradient\\ Filter\\ R4\end{tabular}} & \multicolumn{1}{c|}{1}        & 76.5        & 2.02M   \\
                                                                                       & \multicolumn{1}{c|}{2}        & 92.5        & \multicolumn{1}{c|}{7.93M}   &                                                                                        & \multicolumn{1}{c|}{2}        & 95.6        & 7.93M   &                                                                                        & \multicolumn{1}{c|}{2}        & 75.1        & \multicolumn{1}{c|}{8.07M}   &                                                                                        & \multicolumn{1}{c|}{2}        & 79.5        & 8.07M   \\
                                                                                       & \multicolumn{1}{c|}{3}        & 92.7        & \multicolumn{1}{c|}{13.99M}  &                                                                                        & \multicolumn{1}{c|}{3}        & 95.6        & 13.98M  &                                                                                        & \multicolumn{1}{c|}{3}        & 75.4        & \multicolumn{1}{c|}{14.12M}  &                                                                                        & \multicolumn{1}{c|}{3}        & 79.5        & 14.12M  \\
                                                                                       & \multicolumn{1}{c|}{4}        & 93.4        & \multicolumn{1}{c|}{19.12M}  &                                                                                        & \multicolumn{1}{c|}{4}        & 95.6        & 20.04M  &                                                                                        & \multicolumn{1}{c|}{4}        & 76.1        & \multicolumn{1}{c|}{19.26M}  &                                                                                        & \multicolumn{1}{c|}{4}        & 80.5        & 20.17M  \\ \hline
\multirow{4}{*}{\begin{tabular}[c]{@{}c@{}}PTQ\\ +Gradient\\ Filter\\ R7\end{tabular}} & \multicolumn{1}{c|}{1}        & 91.2        & \multicolumn{1}{c|}{303.10K} & \multirow{4}{*}{\begin{tabular}[c]{@{}c@{}}PTQ\\ +Gradient\\ Filter\\ R7\end{tabular}} & \multicolumn{1}{c|}{1}        & 93.6        & 303.10K & \multirow{4}{*}{\begin{tabular}[c]{@{}c@{}}PTQ\\ +Gradient\\ Filter\\ R7\end{tabular}} & \multicolumn{1}{c|}{1}        & 73.7        & \multicolumn{1}{c|}{441.34K} & \multirow{4}{*}{\begin{tabular}[c]{@{}c@{}}PTQ\\ +Gradient\\ Filter\\ R7\end{tabular}} & \multicolumn{1}{c|}{1}        & 76.5        & 441.34K \\
                                                                                       & \multicolumn{1}{c|}{2}        & 91.5        & \multicolumn{1}{c|}{3.21M}   &                                                                                        & \multicolumn{1}{c|}{2}        & 95.5        & 3.21M   &                                                                                        & \multicolumn{1}{c|}{2}        & 74.5        & \multicolumn{1}{c|}{3.35M}   &                                                                                        & \multicolumn{1}{c|}{2}        & 79.4        & 3.35M   \\
                                                                                       & \multicolumn{1}{c|}{3}        & 91.6        & \multicolumn{1}{c|}{6.12M}   &                                                                                        & \multicolumn{1}{c|}{3}        & 95.4        & 6.12M   &                                                                                        & \multicolumn{1}{c|}{3}        & 74.5        & \multicolumn{1}{c|}{6.26M}   &                                                                                        & \multicolumn{1}{c|}{3}        & 79.5        & 6.26M   \\
                                                                                       & \multicolumn{1}{c|}{4}        & 92.6        & \multicolumn{1}{c|}{8.90M}   &                                                                                        & \multicolumn{1}{c|}{4}        & 95.5        & 9.03M   &                                                                                        & \multicolumn{1}{c|}{4}        & 75.3        & \multicolumn{1}{c|}{9.04M}   &                                                                                        & \multicolumn{1}{c|}{4}        & 79.6        & 9.17M   \\ \hline
\multirow{4}{*}{PSQ}                                                                   & \multicolumn{1}{c|}{1}        & 91.4        & \multicolumn{1}{c|}{128.25M} & \multirow{4}{*}{PSQ}                                                                   & \multicolumn{1}{c|}{1}        & 93.6        & 128.25M & \multirow{4}{*}{PSQ}                                                                   & \multicolumn{1}{c|}{1}        & 73.9        & \multicolumn{1}{c|}{128.39M} & \multirow{4}{*}{PSQ}                                                                   & \multicolumn{1}{c|}{1}        & 76.4        & 128.39M \\
                                                                                       & \multicolumn{1}{c|}{2}        & 93.3        & \multicolumn{1}{c|}{487.68M} &                                                                                        & \multicolumn{1}{c|}{2}        & 96.1        & 487.68M &                                                                                        & \multicolumn{1}{c|}{2}        & 77.7        & \multicolumn{1}{c|}{487.82M} &                                                                                        & \multicolumn{1}{c|}{2}        & 80.3        & 487.82M \\
                                                                                       & \multicolumn{1}{c|}{3}        & 93.4        & \multicolumn{1}{c|}{847.15M} &                                                                                        & \multicolumn{1}{c|}{3}        & 96.2        & 847.13M &                                                                                        & \multicolumn{1}{c|}{3}        & 77.9        & \multicolumn{1}{c|}{847.29M} &                                                                                        & \multicolumn{1}{c|}{3}        & 80.5        & 847.27M \\
                                                                                       & \multicolumn{1}{c|}{4}        & 94.5        & \multicolumn{1}{c|}{1.14G}   &                                                                                        & \multicolumn{1}{c|}{4}        & 96.4        & 1.21G   &                                                                                        & \multicolumn{1}{c|}{4}        & 78.0        & \multicolumn{1}{c|}{1.14G}   &                                                                                        & \multicolumn{1}{c|}{4}        & 82.2        & 1.21G   \\ \hline
\multirow{4}{*}{\begin{tabular}[c]{@{}c@{}}PSQ\\ +Gradient\\ Filter\\ R2\end{tabular}} & \multicolumn{1}{c|}{1}        & 91.3        & \multicolumn{1}{c|}{8.18M}   & \multirow{4}{*}{\begin{tabular}[c]{@{}c@{}}PSQ\\ +Gradient\\ Filter\\ R2\end{tabular}} & \multicolumn{1}{c|}{1}        & 93.5        & 8.18M   & \multirow{4}{*}{\begin{tabular}[c]{@{}c@{}}PSQ\\ +Gradient\\ Filter\\ R2\end{tabular}} & \multicolumn{1}{c|}{1}        & 73.8        & \multicolumn{1}{c|}{8.31M}   & \multirow{4}{*}{\begin{tabular}[c]{@{}c@{}}PSQ\\ +Gradient\\ Filter\\ R2\end{tabular}} & \multicolumn{1}{c|}{1}        & 76.4        & 8.31M   \\
                                                                                       & \multicolumn{1}{c|}{2}        & 92.6        & \multicolumn{1}{c|}{26.80M}  &                                                                                        & \multicolumn{1}{c|}{2}        & 96.0        & 26.80M  &                                                                                        & \multicolumn{1}{c|}{2}        & 76.0        & \multicolumn{1}{c|}{26.94M}  &                                                                                        & \multicolumn{1}{c|}{2}        & 80.1        & 26.94M  \\
                                                                                       & \multicolumn{1}{c|}{3}        & 92.8        & \multicolumn{1}{c|}{45.45M}  &                                                                                        & \multicolumn{1}{c|}{3}        & 96.1        & 45.44M  &                                                                                        & \multicolumn{1}{c|}{3}        & 75.9        & \multicolumn{1}{c|}{45.59M}  &                                                                                        & \multicolumn{1}{c|}{3}        & 80.0        & 45.58M  \\
                                                                                       & \multicolumn{1}{c|}{4}        & 93.7        & \multicolumn{1}{c|}{60.01M}  &                                                                                        & \multicolumn{1}{c|}{4}        & 96.1        & 64.07M  &                                                                                        & \multicolumn{1}{c|}{4}        & 76.3        & \multicolumn{1}{c|}{60.15M}  &                                                                                        & \multicolumn{1}{c|}{4}        & 80.9        & 64.21M  \\ \hline
\multirow{4}{*}{\begin{tabular}[c]{@{}c@{}}PSQ\\ +Gradient\\ Filter\\ R4\end{tabular}} & \multicolumn{1}{c|}{1}        & 91.4        & \multicolumn{1}{c|}{1.88M}   & \multirow{4}{*}{\begin{tabular}[c]{@{}c@{}}PSQ\\ +Gradient\\ Filter\\ R4\end{tabular}} & \multicolumn{1}{c|}{1}        & 93.6        & 1.88M   & \multirow{4}{*}{\begin{tabular}[c]{@{}c@{}}PSQ\\ +Gradient\\ Filter\\ R4\end{tabular}} & \multicolumn{1}{c|}{1}        & 73.5        & \multicolumn{1}{c|}{2.02M}   & \multirow{4}{*}{\begin{tabular}[c]{@{}c@{}}PSQ\\ +Gradient\\ Filter\\ R4\end{tabular}} & \multicolumn{1}{c|}{1}        & 76.5        & 2.02M   \\
                                                                                       & \multicolumn{1}{c|}{2}        & 92.6        & \multicolumn{1}{c|}{7.93M}   &                                                                                        & \multicolumn{1}{c|}{2}        & 95.6        & 7.93M   &                                                                                        & \multicolumn{1}{c|}{2}        & 75.3        & \multicolumn{1}{c|}{8.07M}   &                                                                                        & \multicolumn{1}{c|}{2}        & 79.5        & 8.07M   \\
                                                                                       & \multicolumn{1}{c|}{3}        & 92.7        & \multicolumn{1}{c|}{13.99M}  &                                                                                        & \multicolumn{1}{c|}{3}        & 95.6        & 13.98M  &                                                                                        & \multicolumn{1}{c|}{3}        & 75.1        & \multicolumn{1}{c|}{14.12M}  &                                                                                        & \multicolumn{1}{c|}{3}        & 79.6        & 14.12M  \\
                                                                                       & \multicolumn{1}{c|}{4}        & 93.2        & \multicolumn{1}{c|}{19.12M}  &                                                                                        & \multicolumn{1}{c|}{4}        & 95.5        & 20.04M  &                                                                                        & \multicolumn{1}{c|}{4}        & 76.2        & \multicolumn{1}{c|}{19.26M}  &                                                                                        & \multicolumn{1}{c|}{4}        & 80.2        & 20.17M  \\ \hline
\multirow{4}{*}{\begin{tabular}[c]{@{}c@{}}PSQ\\ +Gradient\\ Filter\\ R7\end{tabular}} & \multicolumn{1}{c|}{1}        & 91.2        & \multicolumn{1}{c|}{303.10K} & \multirow{4}{*}{\begin{tabular}[c]{@{}c@{}}PSQ\\ +Gradient\\ Filter\\ R7\end{tabular}} & \multicolumn{1}{c|}{1}        & 93.6        & 303.10K & \multirow{4}{*}{\begin{tabular}[c]{@{}c@{}}PSQ\\ +Gradient\\ Filter\\ R7\end{tabular}} & \multicolumn{1}{c|}{1}        & 73.5        & \multicolumn{1}{c|}{441.34K} & \multirow{4}{*}{\begin{tabular}[c]{@{}c@{}}PSQ\\ +Gradient\\ Filter\\ R7\end{tabular}} & \multicolumn{1}{c|}{1}        & 76.5        & 441.34K \\
                                                                                       & \multicolumn{1}{c|}{2}        & 91.4        & \multicolumn{1}{c|}{3.21M}   &                                                                                        & \multicolumn{1}{c|}{2}        & 95.5        & 3.21M   &                                                                                        & \multicolumn{1}{c|}{2}        & 74.4        & \multicolumn{1}{c|}{3.35M}   &                                                                                        & \multicolumn{1}{c|}{2}        & 79.5        & 3.35M   \\
                                                                                       & \multicolumn{1}{c|}{3}        & 91.6        & \multicolumn{1}{c|}{6.12M}   &                                                                                        & \multicolumn{1}{c|}{3}        & 95.4        & 6.12M   &                                                                                        & \multicolumn{1}{c|}{3}        & 74.5        & \multicolumn{1}{c|}{6.26M}   &                                                                                        & \multicolumn{1}{c|}{3}        & 79.6        & 6.26M   \\
                                                                                       & \multicolumn{1}{c|}{4}        & 92.7        & \multicolumn{1}{c|}{8.90M}   &                                                                                        & \multicolumn{1}{c|}{4}        & 95.5        & 9.03M   &                                                                                        & \multicolumn{1}{c|}{4}        & 75.5        & \multicolumn{1}{c|}{9.04M}   &                                                                                        & \multicolumn{1}{c|}{4}        & 79.6        & 9.17M  
                                                                                       \\ \bottomrule
\end{tabular}%
}
\caption{Experimental results for ResNet18 and ResNet34 with different gradient quantization methods (\textit{i.e.}, PTQ~\cite{banner2018scalable} and PSQ~\cite{chen2020statistical}) and hyper-parameter selections on CIFAR10/100. Feature map, activation, weight and gradient are quantized to INT8. ``ACC'' is short for accuracy. ``\#Layers'' is short for ``the number of \textit{active} convolution layers''. For example. \#Layers equals to 2 means that the last two convolutional layers are trained. ``Gradient Filter R2/4/7'' use proposed gradient filtering method with patch size $2\times 2$, $4\times 4$ and $7\times 7$, respectively.}
\label{tab:cifar_gq}
\end{table*}

\end{document}